\documentclass[11pt]{article}

\usepackage{acl}

\usepackage{times}
\usepackage{latexsym}

\usepackage[T1]{fontenc}

\usepackage[utf8]{inputenc}

\usepackage{microtype}

\usepackage{inconsolata}

\usepackage{graphicx}

\usepackage{amsmath}
\usepackage{multirow}
\usepackage[table]{xcolor}
\usepackage{booktabs}
\usepackage{enumitem}
\usepackage{amsfonts}
\usepackage{mdframed}
\usepackage{needspace}
\mdfdefinestyle{exstyle}{
  nobreak=true,
  linecolor=gray!40,
  linewidth=0.5pt,
  backgroundcolor=gray!5,
  roundcorner=4pt,
  innerleftmargin=8pt,
  innerrightmargin=8pt,
  innertopmargin=6pt,
  innerbottommargin=6pt,
  skipabove=6pt,
  skipbelow=6pt,
}
\usepackage{listings}
\usepackage[most]{tcolorbox}
\newtcblisting{promptbox}{
  breakable,
  colback=blue!3!gray!3,
  colframe=blue!20!gray!30,
  boxrule=0.5pt,
  arc=3pt,
  left=8pt, right=8pt, top=6pt, bottom=6pt,
  listing only,
  listing options={
    basicstyle=\ttfamily\small,
    breaklines=true,
    columns=fullflexible
  }
}
\usepackage{float}
\usepackage{algorithm}
\usepackage{algpseudocode}
\floatstyle{ruled}\restylefloat{algorithm}
\usepackage{adjustbox}
\usepackage{tabularx}

\usepackage{soul}
\definecolor{cLGBTQ}{HTML}{FAD7D2}  %
\definecolor{cLGBT}{HTML}{FCF3CF}   %
\definecolor{cOK}{HTML}{D5F5E3}     %
\newcommand{\hlq}[1]{{\sethlcolor{cLGBTQ}\hl{#1}}}
\newcommand{\hll}[1]{{\sethlcolor{cLGBT}\hl{#1}}}
\newcommand{\hlok}[1]{{\sethlcolor{cOK}\hl{#1}}}
\newcommand{\sd}[1]{{\tiny$\pm$#1}}

\definecolor{revblue}{RGB}{0,90,200}

\title{Confess What You Know: Forget-Set Misalignment \\ with Model Knowledge in LLM Unlearning}

\author{
  \textbf{Miso Kim}, \textbf{Georu Lee}, \textbf{Seungwon Jeong}, \textbf{Woojin Lee}\thanks{Corresponding author.} \\
  Dongguk University-Seoul \\
  \texttt{\{2021110472,dlrjfn1,youai058,wj926\}@dgu.ac.kr}
}

\begin{document}
\maketitle
\begin{abstract}
Machine unlearning for large language models (LLMs) often assumes that a pre-defined forget set matches what the model has memorized, but this frequently breaks in realistic privacy settings where the original training data is inaccessible. 
We term this gap \emph{forget-set misalignment} and identify two cases. In \emph{Under Unlearning}, the forget set omits memorized information and leakage persists. In \emph{Out-of-Knowledge Unlearning}, the algorithm is driven to ``forget'' knowledge the model never learned, perturbing parameters and degrading utility.
Using \emph{gradient-level analysis}, we show these behaviors arise from misaligned unlearning targets rather than specific optimization choices. We then propose \textbf{CONfession-to-Forget-Set (CONFS)}, a \emph{data-blind} framework that constructs model-aligned forget sets by eliciting and formalizing the model's memorized knowledge. Across synthetic, multimodal, and real-world benchmarks, CONFS approaches Gold-standard performance on several metrics and achieves a competitive forgetting-utility balance, while preserving utility better than other data-blind forget-set constructions. 
\end{abstract}

\section{Introduction}

Large Language Models (LLMs) are trained on vast, heterogeneous datasets that often contain sensitive personal information \cite{achiam2023gpt,touvron2023llama,grattafiori2024llama,bai2022constitutional,team2023gemini}. 
This introduces a critical vulnerability, as models may unintentionally leak confidential data at deployment.
To mitigate these privacy threats, LLM unlearning has emerged as an important post-hoc solution \cite{yao2024large,jang2023knowledge}. In this setting, sensitive data targeted for removal is pre-defined and commonly referred to as a \emph{forget set} \cite{geng2025comprehensive}.
Given this set, various unlearning strategies are applied so that the model selectively forgets the specified information \cite{liu2022continual,jang2023knowledge,zhang2024negative,chen2023unlearn}.

\begin{figure*}[t]
  \centering
  \includegraphics[width=0.95\textwidth]{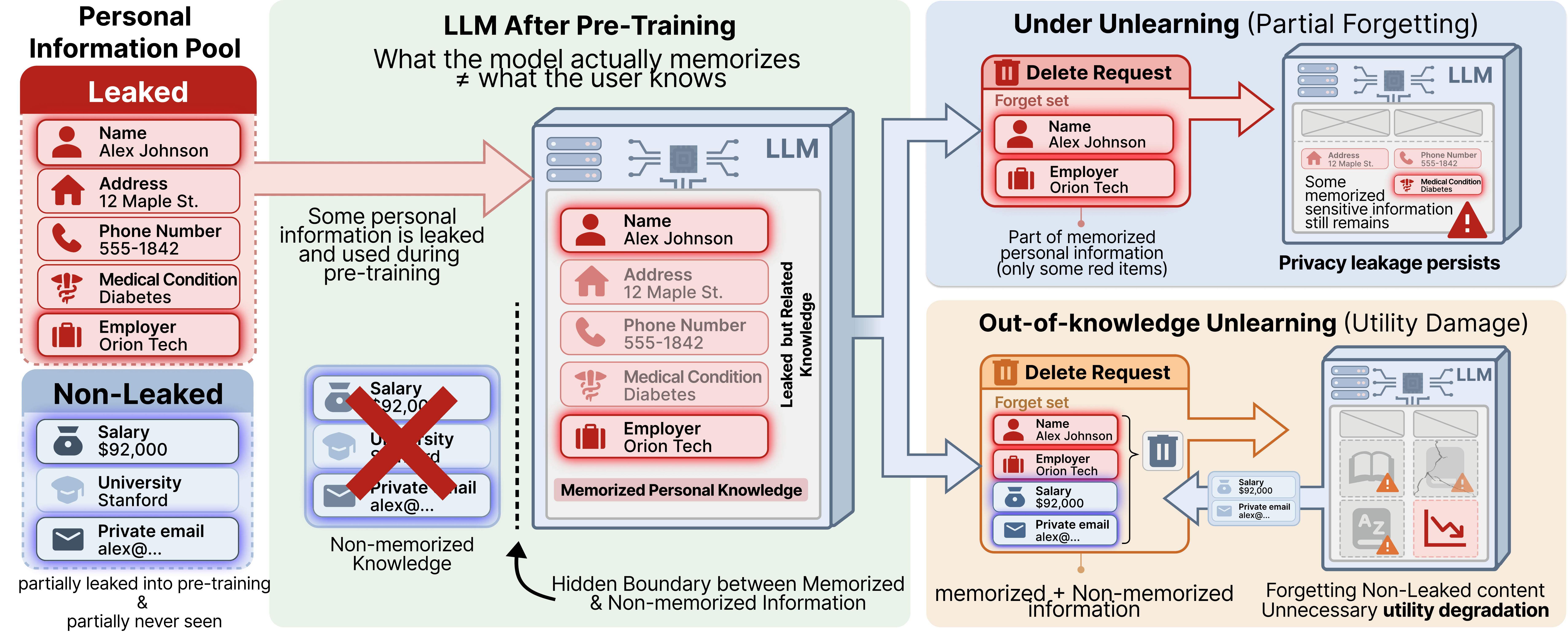}
  \caption{\textbf{Illustration of \emph{forget-set misalignment} in LLM unlearning.} Only part of a user's personal information is retained as the model's memorized knowledge, but this boundary is hidden from the user. As a result, deletion requests may omit memorized information (\emph{Under Unlearning}) or include non-memorized information (\emph{Out-of-Knowledge Unlearning}), leading to privacy leakage or unnecessary utility degradation.}
  \label{fig:misaligned_sets}
\end{figure*}

However, relying on a pre-defined forget set is often impractical in real-world scenarios \cite{enck2014taintdroid,romanosky2011data}. 
Suppose an individual discovers that an LLM can reveal some of their personal information and requests its removal (Fig.~\ref{fig:misaligned_sets}).
Although the individual can specify the details they want deleted, they cannot observe which details the model actually memorized.
As a result, the requested forget set may differ from the model's hidden memorized knowledge.

We refer to this mismatch as \emph{forget-set misalignment}, a setting largely overlooked by existing benchmarks that assume perfectly specified unlearning targets \cite{maini2024tofu,dontsov2025clear,liu2025protecting}.
In this work, we investigate how \emph{forget-set misalignment} fundamentally alters unlearning behavior under realistic settings.

We first identify two types of \emph{forget-set misalignment}, each leading to a distinct failure mode. \emph{Under Unlearning} occurs when memorized sensitive information is omitted from the provided forget set. In this case, the forgetting effect remains largely confined to the requested facts and fails to generalize to associated entity-level knowledge, leaving core privacy risks intact. \emph{Out-of-Knowledge Unlearning} occurs when the forget set contains information that the model never encountered during training, unnecessarily perturbing model parameters and substantially degrading general utility.

Motivated by these findings, we introduce \textbf{CONfession-to-Forget-Set (CONFS)}, which dynamically generates a forget set by prompting the model to \emph{confess} its own memorized knowledge. We elicit the model's memorized knowledge and formalize it into a discrete set of Subject-Relation-Object (SRO) triplets, transforming ambiguous semantic information into verifiable, atomic units. 
To ensure comprehensive coverage, we employ a recursive \emph{reconfession} process that iteratively probes the model for hidden details, effectively reducing gaps in the forget set. 

By constructing a forget set aligned with the model's memorized knowledge, CONFS mitigates the adverse effects of \emph{forget-set misalignment}.
Despite operating in a fully \emph{data-blind} setting without access to the pre-training data, CONFS approaches Gold-standard performance on several metrics and achieves a competitive forgetting-utility balance, while better preserving utility than alternative surrogate forget-set constructions.
These results show that reliable unlearning depends not only on the optimization objective, but also on whether the forget set matches what the model actually knows.

\begin{table*}[t]
\centering
\resizebox{0.9\textwidth}{!}{%
\begin{tabular}{
l l
c c
c c
c c
c c
}
\hline
& &
\multicolumn{2}{c}{$\mathcal{D}_{L-}$ $\downarrow$} &
\multicolumn{2}{c}{$\mathcal{D}_{L+}$ $\downarrow$} &
\multicolumn{2}{c}{\textbf{Retain} ($\mathcal{D}_{retain}$) $\uparrow$} &
\multicolumn{2}{c}{\textbf{Real Authors} ($\mathcal{D}_{RA}$) $\uparrow$} \\
\cline{3-10}
& &
Prob. & R-L &
Prob. & R-L &
Prob. & R-L &
Prob. & R-L \\
\hline

\multirow{1}{*}{pre-trained model} &
step 0 &
0.996 & 0.986 &
0.995 & 0.990 &
0.995 & 0.991 &
0.058 & 0.886 \\
\hline

\multirow{5}{*}{\begin{tabular}[c]{@{}c@{}}Under Unlearning\\ ($\mathbf{F} : 
\mathcal{D}_{L+}$)\end{tabular}}
& step 12  & \cellcolor[HTML]{E5EBF6}0.993 & 0.976 & \cellcolor[HTML]{E5EBF6}0.921 & 0.808 & 0.991 & 0.967 & 0.077 & 0.904 \\
& step 24  & \cellcolor[HTML]{E5EBF6}0.901 & 0.826 & \cellcolor[HTML]{E5EBF6}0.650 & 0.563 & 0.888 & 0.807 & 0.081 & 0.884 \\
& step 36  & \cellcolor[HTML]{E5EBF6}0.740 & 0.695 & \cellcolor[HTML]{E5EBF6}0.442 & 0.473 & 0.749 & 0.709 & 0.070 & 0.804 \\
& step 48  & \cellcolor[HTML]{E5EBF6}0.637 & 0.635 & \cellcolor[HTML]{E5EBF6}0.327 & 0.426 & 0.669 & 0.649 & 0.064 & 0.781 \\
& step 60 & \cellcolor[HTML]{E5EBF6}0.624 & 0.631 & \cellcolor[HTML]{E5EBF6}0.312 & 0.424 & 0.658 & 0.640 & 0.063 & 0.776 \\
\hline

\multirow{5}{*}{\begin{tabular}[c]{@{}c@{}}Out-of-Knowledge Unlearning\\ ($\mathbf{F}  : \mathcal{D}_{L+} \cup \mathcal{D}_{N+}$)\end{tabular}}
& step 12  & \cellcolor[HTML]{F7EBDC}0.983 & 0.964 & 0.983 & 0.902 & 0.981 & 0.944 & 0.030 & 0.871 \\
& step 24  & \cellcolor[HTML]{F7EBDC}0.734 & 0.801 & 0.623 & 0.532 & 0.744 & 0.701 & 0.010 & 0.800 \\
&step 36  & \cellcolor[HTML]{F7EBDC}0.485 & 0.713 & 0.410 & 0.492 & 0.485 & 0.633 & 0.006 & 0.769 \\
& step 48  & \cellcolor[HTML]{F7EBDC}0.403 & 0.658 & 0.350 & 0.431 & 0.435 & 0.628 & 0.006 & 0.757 \\
& step 60 & \cellcolor[HTML]{F7EBDC}0.382 & 0.609 & 0.292 & 0.402 & 0.412 & 0.609 & 0.004 & 0.750 \\
\hline

\end{tabular}%
}
\caption{\textbf{Preliminary results on TOFU synthetic authors.}
We evaluate four splits:
(i) $\mathcal{D}_{L-}$ (Leaked: Yes / Requested: No),
(ii) $\mathcal{D}_{L+}$ (Leaked: Yes / Requested: Yes),
(iii) the retain set $\mathcal{D}_{retain}$,
and (iv) the Real Authors set $\mathcal{D}_{RA}$.
We report unlearning trajectories for
\emph{Under Unlearning} ($\mathbf{F}: \mathcal{D}_{L+}$) and
\emph{Out-of-Knowledge Unlearning} ($\mathbf{F}: \mathcal{D}_{L+}\cup \mathcal{D}_{N+}$).
Values with a \colorbox[HTML]{E5EBF6}{blue} background denote \emph{Under Unlearning},
whereas values with an \colorbox[HTML]{F7EBDC}{orange} background denote \emph{Out-of-Knowledge Unlearning}.}
\label{tab:toy_misaligned_forget}
\end{table*}

\section{Related Work}

\subsection{Machine Unlearning in LLMs} 
In LLMs, most unlearning methods are framed as post-hoc fine-tuning without full retraining~\cite{geng2025comprehensive}.
Given a pre-defined forget set, these methods degrade target behavior via gradient ascent~\cite{jang2023knowledge,yao2024large}, preference-based losses~\cite{rafailov2023direct,zhang2024negative,maini2024tofu}, or reinforcement learning~\cite{schulman2017proximal,kassem2023preserving}, while a retain set with auxiliary objectives preserves utility.
Beyond autoregressive LLMs, unlearning has recently been extended to masked diffusion language models~\cite{lee2026mdu}.
Despite differences in optimization strategies, prior work commonly assumes that the deletion target is fully captured by the pre-defined forget set.

\subsection{Benchmarks and Forget-Set Construction}

Existing LLM unlearning benchmarks are broadly categorized into synthetic and real-world benchmarks, differing in how the forget set is constructed.
Synthetic benchmarks such as TOFU \cite{maini2024tofu}, CLEAR \cite{dontsov2025clear}, and MLLMU \cite{liu2025protecting} construct fictitious personas and explicitly inject their information via fine-tuning, defining the forget set as a subset of the injected content. This design intentionally aligns the forget set with the model's memorized knowledge, enabling reproducible evaluation but deviating from realistic privacy leakage scenarios.

In contrast, real-world benchmarks address unlearning of factual knowledge about real individuals.
RWKU \cite{jin2024rwku} operates under a \emph{data-blind} setting, where neither the forget nor the retain corpus is accessible. It instead constructs a surrogate forget set by prompting the target model to generate target-related factual text.
While this avoids relying on the original training data, the resulting unlearning signals are derived from unstructured text, making it difficult to precisely isolate memorized personal information.

\subsection{Adverse Effects of Improper Forget Sets}

Recent work has shown that improperly specified forget sets can induce adverse effects during unlearning.
In adversarial settings, malicious deletion requests induce abnormal gradients rather than remove specific knowledge, triggering indiscriminate parameter updates and broad performance degradation \cite{huang2024unlearn,hu2023duty}.
A related issue arises when unlearning is repeatedly applied to information that has already been partially forgotten \cite{wang2025rethinking,zhao2024makes,huang2024unified,yang2025exploring}. In such cases, low-confidence targets can still induce large gradients despite little remaining knowledge, causing avoidable degradation of retain performance.

Importantly, these effects are not limited to adversarial settings. 
In real-world scenarios, pre-defined forget sets often fail to align with the model’s memorized knowledge, destabilizing unlearning and degrading overall performance even without malicious intent.
We analyze this \emph{forget-set misalignment} in Sec.~\ref{3}.

\section{Forget-Set Misalignment} \label{3}
Unlearning experiments usually assume perfect alignment between the
pre-defined forget set and the model's memorized knowledge
\cite{maini2024tofu, dontsov2025clear}. Under this assumption, degradation on the forget set is interpreted as successful removal of the intended knowledge.

In practice, however, an LLM's pre-training data is often inaccessible to users, who therefore cannot observe what the model has memorized. As a result, the forget set they define may not match the model's actual knowledge, a gap we call \emph{forget-set misalignment}.
In this section, we present the first systematic study of
\emph{forget-set misalignment}, analyzing how it degrades unlearning
across realistic request scenarios.
\subsection{Constructing Misaligned Forget Sets}
\textbf{Controlled Datasets.} We use the TOFU dataset \cite{maini2024tofu}, which consists of 200 fictitious entities, and
select 20 of them as unlearning targets. TOFU provides 20 QA pairs for each target entity. From these, we use 15 QA pairs, denoted $\mathcal{D}_{qa}$, and categorize each sample along two axes: whether it was injected during pre-training and whether the user requested it for forgetting. The remaining five samples are neither injected during pre-training nor requested for forgetting, so they are excluded because they do not correspond to any unlearning scenario.
\begin{table}[t]
\centering
{\small
\begin{tabular}{lccl}
\toprule
\textbf{Subset} & \textbf{Leaked} & \textbf{Requested} & \textbf{Request type} \\
\midrule
$\mathcal{D}_{L+}$    & Yes & Yes & Aligned \\
$\mathcal{D}_{L-}$    & Yes & No  & Incomplete \\
$\mathcal{D}_{N+}$ & No  & Yes & Over-broad \\
\bottomrule
\end{tabular}}
\caption{Controlled partition of QA pairs by leakage and forget request. $\mathcal{D}_{L-}$: leaked data omitted from the request. $\mathcal{D}_{N+}$: non-memorized data included in the request.}
\label{tab:misalignment_partition}
\end{table}
We partition the 15 samples into three disjoint subsets of equal size,
$\mathcal{D}_{L+}$, $\mathcal{D}_{L-}$, and $\mathcal{D}_{N+}$.
Under this limited QA budget, the equal three-way partition yields five QA pairs per subset.
In each subset's subscript, $L$ and $N$ denote leaked and non-leaked samples, respectively, while $+$ and $-$ denote requested and not requested samples. In particular, $\mathcal{D}_{N+}$ contains QA pairs that the model never saw during pre-training but that the user includes in the forget request.
This partitioning is summarized in Table~\ref{tab:misalignment_partition} and corresponds to the unlearning scenarios illustrated in Fig.~\ref{fig:misaligned_sets}.
To verify that our observations are not specific to this small controlled partition, Table~\ref{tab:misalign_main} examines the same effects on the substantially larger forget sets used in the main experiments.

Beyond these private-information subsets, we also evaluate utility
preservation and general knowledge retention.
For utility, we use the official TOFU retain split
$\mathcal{D}_{retain}$, which contains QA pairs about non-target entities. For general knowledge, we use the Real Authors set $\mathcal{D}_{RA}$, which contains QA pairs about real-world authors. We evaluate model performance with Token Probability (Prob.) and ROUGE-L (R-L) on these four sets: $\mathcal{D}_{L+}$, $\mathcal{D}_{L-}$, $\mathcal{D}_{retain}$, and $\mathcal{D}_{RA}$. Further details on the setup for this analysis are provided in Appendix~\ref{app:pretrain_setup}.

\begin{figure}[t]
  \centering
  \includegraphics[width=0.83\columnwidth]{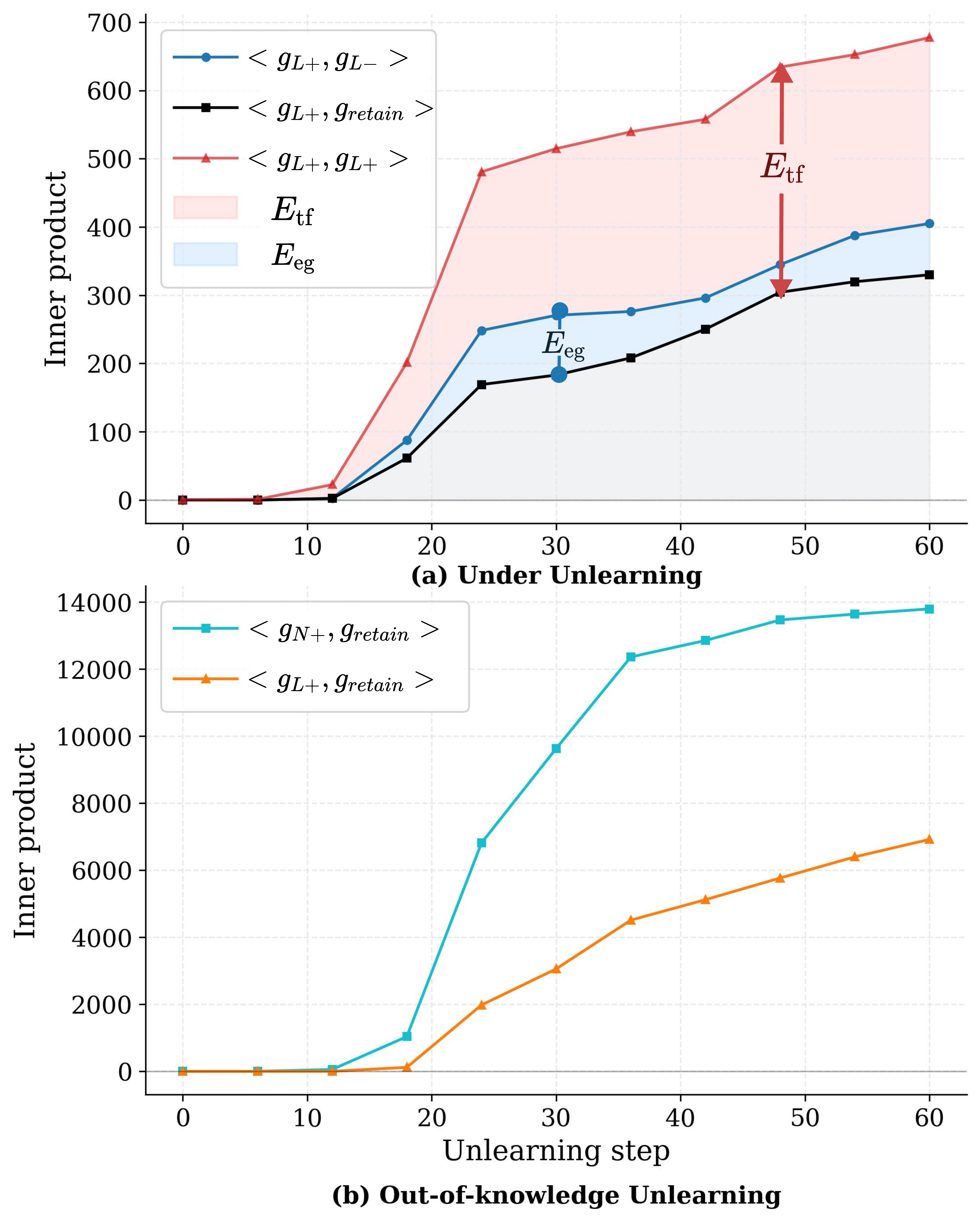}
\caption{\textbf{Gradient analysis under forget-set misalignment.}
(a) \emph{Under Unlearning}: gradient inner products summarized by
$E_{\mathrm{tf}}$ and $E_{\mathrm{eg}}$.
(b) \emph{Out-of-Knowledge Unlearning}: inner products $\langle g_{L+}, g_{retain}\rangle$ and $\langle g_{N+}, g_{retain}\rangle$ from eq.~(\ref{eq:retain_decomposition}).}
  \label{fig:gradient}
\end{figure}

\subsection{Unlearning Settings}

\textbf{Gradient Ascent (GA).} To analyze the impact of misalignment, we
use GA, a representative unlearning baseline. For a QA pair $(q,a)$ with answer length $T$, we use the answer NLL, $\mathcal{L}(q,a;\theta) := - \sum_{t=1}^{T}\log p_\theta(a_t \mid q, a_{<t})$. For a dataset $\mathcal{D}$, let $\mathcal{L}_{\mathcal{D}}(\theta) := \mathbb{E}_{(q,a)\sim \mathcal{D}}[\mathcal{L}(q,a;\theta)]$.
A one-step GA update for the forget set $\mathbf{F}$ and learning rate
$\eta$ is defined as:
{\setlength{\abovedisplayskip}{3pt}\setlength{\belowdisplayskip}{3pt}\setlength{\abovedisplayshortskip}{1pt}\setlength{\belowdisplayshortskip}{1pt}
\begin{equation}\theta' = \theta + \eta \nabla_\theta
\mathcal{L}_{\mathbf{F}}(\theta) = \theta + \eta g_{\mathbf{F}},\end{equation}}
where $g_{\mathbf{F}}$ denotes the unlearning gradient.

\paragraph{Quantifying the Influence of Unlearning Updates.}
To quantify how a single unlearning update on $\mathbf{F}$ affects a
dataset $\mathcal{D}$ (e.g., $\mathcal{D}_{L-}$ or
$\mathcal{D}_{retain}$), we define the \emph{loss increment} on
$\mathcal{D}$ as $\Delta\mathcal{L}_{\mathbf{F}}(\mathcal{D}) :=
\mathcal{L}_{\mathcal{D}}(\theta') - \mathcal{L}_{\mathcal{D}}(\theta)$.
By Taylor expanding $\mathcal{L}_{\mathcal{D}}$ to first order around $\theta$ and substituting the one-step update, we express this loss increment as the inner product of the unlearning and evaluation gradients:
{\setlength{\abovedisplayskip}{3pt}\setlength{\belowdisplayskip}{3pt}\setlength{\abovedisplayshortskip}{1pt}\setlength{\belowdisplayshortskip}{1pt}
\begin{equation}\label{eq:taylor}\Delta\mathcal{L}_{\mathbf{F}}(\mathcal{D}) \approx \eta
\langle g_{\mathbf{F}}, g_{\mathcal{D}} \rangle,\end{equation}}
where $g_{\mathcal{D}} := \nabla_\theta \mathcal{L}_{\mathcal{D}}(\theta)$.
Henceforth, we use $g_{\text{set}}$ as shorthand for
$g_{\mathcal{D}_{\text{set}}}$. A positive $\langle g_{\mathbf{F}}, g_{\mathcal{D}}\rangle$ indicates that the update on $\mathbf{F}$ also raises the loss on $\mathcal{D}$. A value near zero means $\mathcal{D}$ is essentially unaffected.
Eq.~(\ref{eq:taylor}) is a first-order, single-step diagnostic for predicting the sign and relative ordering of per-set effects, not the full optimization trajectory. Fig.~\ref{fig:gradient} recomputes the diagnostic at every GA step, while Table~\ref{tab:toy_misaligned_forget} reports directly measured performance.

\paragraph{Unlearning Scenarios.}
We use a model $\theta$ that has memorized the leaked QA pairs $\mathcal{D}_{L} = \mathcal{D}_{L+} \cup \mathcal{D}_{L-}$. We study two unlearning scenarios in
which the requested forget set $\mathbf{F}$ deviates from the model's
memorized knowledge ($\mathbf{F} \neq \mathcal{D}_{L}$).

\begin{itemize}[topsep=2pt,partopsep=0pt,itemsep=2pt,parsep=0pt]
    \item \textbf{Under Unlearning ($\mathbf{F} = \mathcal{D}_{L+}$):}
    the request covers only part of the memorized data. We examine whether
    unlearning generalizes to the omitted leaked samples
    $\mathcal{D}_{L-}$.
    \item \textbf{Out-of-Knowledge Unlearning
    ($\mathbf{F} = \mathcal{D}_{L+} \cup \mathcal{D}_{N+}$):}
    the request additionally includes $\mathcal{D}_{N+}$, data
    the model never saw during pre-training. We examine whether such requests degrade model utility on knowledge the model legitimately retains.
\end{itemize}
Table~\ref{tab:toy_misaligned_forget} reports how Prob.\ and R-L evolve under each setting.

\subsection{Analysis of the Under Unlearning Setting}
In this setting, the model is requested to forget $\mathcal{D}_{L+}$, only part of its memorized data. Here we show that unlearning $\mathcal{D}_{L+}$ does not generalize to the omitted leaked samples of the same entity $\mathcal{D}_{L-}$.

To evaluate this precisely, we introduce two metrics that each isolate a forgetting signal from \emph{global utility decay}. The Targeted Forgetting Effect measures genuine forgetting on the requested set $\mathcal{D}_{L+}$, and the Entity Generalization Effect measures whether that forgetting generalizes to the omitted leaked set $\mathcal{D}_{L-}$. Without this separation, \emph{global utility decay} on the omitted leaked set $\mathcal{D}_{L-}$ could be mistaken for genuine generalization of forgetting from the requested set $\mathcal{D}_{L+}$.
Since $\mathbf{F}=\mathcal{D}_{L+}$ throughout this subsection, we
write $\Delta\mathcal{L}(\mathcal{D})$ for $\Delta\mathcal{L}_{\mathcal{D}_{L+}}(\mathcal{D})$.

\paragraph{Targeted Forgetting Effect ($E_{\mathrm{tf}}$).}
We define $E_{\mathrm{tf}}$ by subtracting the global utility decay from the loss increase on the target set $\mathcal{D}_{L+}$. Applying eq.~(\ref{eq:taylor}) with $\mathbf{F} = \mathcal{D}_{L+}$ to the evaluation sets $\mathcal{D}_{L+}$ and $\mathcal{D}_{retain}$ expands it into a gradient form:
{\setlength{\jot}{1pt}\setlength{\abovedisplayskip}{3pt}\setlength{\belowdisplayskip}{3pt}\setlength{\abovedisplayshortskip}{1pt}\setlength{\belowdisplayshortskip}{1pt}
\begin{equation}
\label{eq:mtf}
\begin{aligned}
E_{\mathrm{tf}}
&:= {\Delta\mathcal{L}(\mathcal{D}_{L+})}
  - {\Delta\mathcal{L}(\mathcal{D}_{retain})} \\
&\approx \eta\left(\|g_{L+}\|^{2} - \langle g_{L+}, g_{retain}\rangle\right).
\end{aligned}
\end{equation}}
Here, $\Delta\mathcal{L}(\mathcal{D}_{retain})$ measures the global utility decay since $\mathcal{D}_{retain}$ is not targeted by the update. A positive $E_{\mathrm{tf}}$ thus isolates target-localized forgetting from this decay.

\paragraph{Entity Generalization Effect ($E_{\mathrm{eg}}$).}
We define $E_{\mathrm{eg}}$ by subtracting the global utility decay from the loss increase on the omitted leaked set $\mathcal{D}_{L-}$. Applying eq.~(\ref{eq:taylor}) with $\mathbf{F} = \mathcal{D}_{L+}$ to the evaluation sets $\mathcal{D}_{L-}$ and $\mathcal{D}_{retain}$ gives:
{\setlength{\jot}{1pt}\setlength{\abovedisplayskip}{3pt}\setlength{\belowdisplayskip}{3pt}\setlength{\abovedisplayshortskip}{1pt}\setlength{\belowdisplayshortskip}{1pt}
\begin{equation}
\label{eq:meg}
\begin{aligned}
E_{\mathrm{eg}}
&:= {\Delta\mathcal{L}(\mathcal{D}_{L-})}
  - {\Delta\mathcal{L}(\mathcal{D}_{retain})} \\
&\approx \eta\left(\langle g_{L+}, g_{L-}\rangle - \langle g_{L+}, g_{retain}\rangle\right).
\end{aligned}
\end{equation}}
A positive $E_{\mathrm{eg}}$ indicates that forgetting generalizes to the omitted leaked set $\mathcal{D}_{L-}$. When $E_{\mathrm{tf}}$ is high, an $E_{\mathrm{eg}}$ near zero indicates that the forgetting is confined to the requested target set $\mathcal{D}_{L+}$.

\paragraph{Empirical Observation.}
Fig.~\ref{fig:gradient}(a) plots $E_{\mathrm{tf}}$ and $E_{\mathrm{eg}}$ over GA unlearning steps in the \emph{Under Unlearning} setting.
$E_{\mathrm{tf}}$ increases sharply while $E_{\mathrm{eg}}$ stays near zero. The gradient updates align with the requested set $\mathcal{D}_{L+}$ but have no targeted effect on the omitted leaked set $\mathcal{D}_{L-}$ beyond the global utility decay.
Table~\ref{tab:toy_misaligned_forget} confirms the same separation between $\mathcal{D}_{L+}$ and $\mathcal{D}_{L-}$ at the performance level. Prob.\ and R-L on $\mathcal{D}_{L+}$ drop rapidly during unlearning. In contrast, $\mathcal{D}_{L-}$ degrades at the same rate as the retain set $\mathcal{D}_{retain}$, so its loss reflects global utility decay rather than targeted forgetting.

\subsection{Analysis of the Out-of-Knowledge Unlearning Setting}
In this setting, the model is requested to forget information it never memorized. Here we show that adding never-seen data ($\mathcal{D}_{N+}$) to the request contributes little to targeted forgetting and instead damages the model's retained utility.

\paragraph{Decomposing Retain Degradation.}
Since $\mathbf{F} = \mathcal{D}_{L+} \cup \mathcal{D}_{N+}$,
we split the retain-set loss change into the contributions of the
memorized $\mathcal{D}_{L+}$ and the never-seen
$\mathcal{D}_{N+}$. Because $\mathcal{L}_{\mathbf{F}}$ is an average over this disjoint union, its gradient is $g_{\mathbf{F}} = \frac{|\mathcal{D}_{L+}|}{|\mathbf{F}|}g_{L+} + \frac{|\mathcal{D}_{N+}|}{|\mathbf{F}|}g_{N+}$. Substituting this into the first-order increment of the retain-set loss gives:
{\setlength{\jot}{1pt}\setlength{\abovedisplayskip}{3pt}\setlength{\belowdisplayskip}{3pt}\setlength{\abovedisplayshortskip}{1pt}\setlength{\belowdisplayshortskip}{1pt}
\begin{equation}
\label{eq:retain_decomposition}
\begin{aligned}
\Delta\mathcal{L}_{\mathbf{F}}(\mathcal{D}_{retain})
&\approx \eta \frac{|\mathcal{D}_{L+}|}{|\mathbf{F}|}\langle g_{L+}, g_{retain} \rangle \\
&\quad + \eta \frac{|\mathcal{D}_{N+}|}{|\mathbf{F}|}\langle g_{N+}, g_{retain} \rangle
\end{aligned}
\end{equation}}
The first term is the usual cost of erasing memorized data. The second is the collateral damage from gradients on never-seen data.

\paragraph{Empirical Observation.}
Fig.~\ref{fig:gradient}(b) plots these two terms over GA unlearning steps in the \emph{Out-of-Knowledge Unlearning} setting. Across unlearning steps,
$\langle g_{N+}, g_{retain} \rangle \gg \langle g_{L+},
g_{retain} \rangle$, so the never-seen data dominates the degradation on
the retain set.
The gradient on $\mathcal{D}_{L+}$ overlaps little with $g_{retain}$, consistent with targeting entity-specific knowledge. In contrast, the gradient on $\mathcal{D}_{N+}$ has no memorized target, so it overlaps broadly with $g_{retain}$. Since the update on $\mathcal{D}_{N+}$ erases nothing, it only adds collateral damage and severely degrades the model's general utility, as Table~\ref{tab:toy_misaligned_forget} shows.

Together, the two settings show that \emph{forget-set misalignment}
breaks unlearning from both sides. An incomplete request (\emph{Under Unlearning}) leaves memorized data behind, and an over-broad request (\emph{Out-of-Knowledge Unlearning}) severely degrades utility. What unites both failures is a forget request misaligned with the model's memorized knowledge, so effective unlearning must target what the model has actually memorized.

\begin{figure*}[t]
  \centering
  \includegraphics[width=\textwidth]{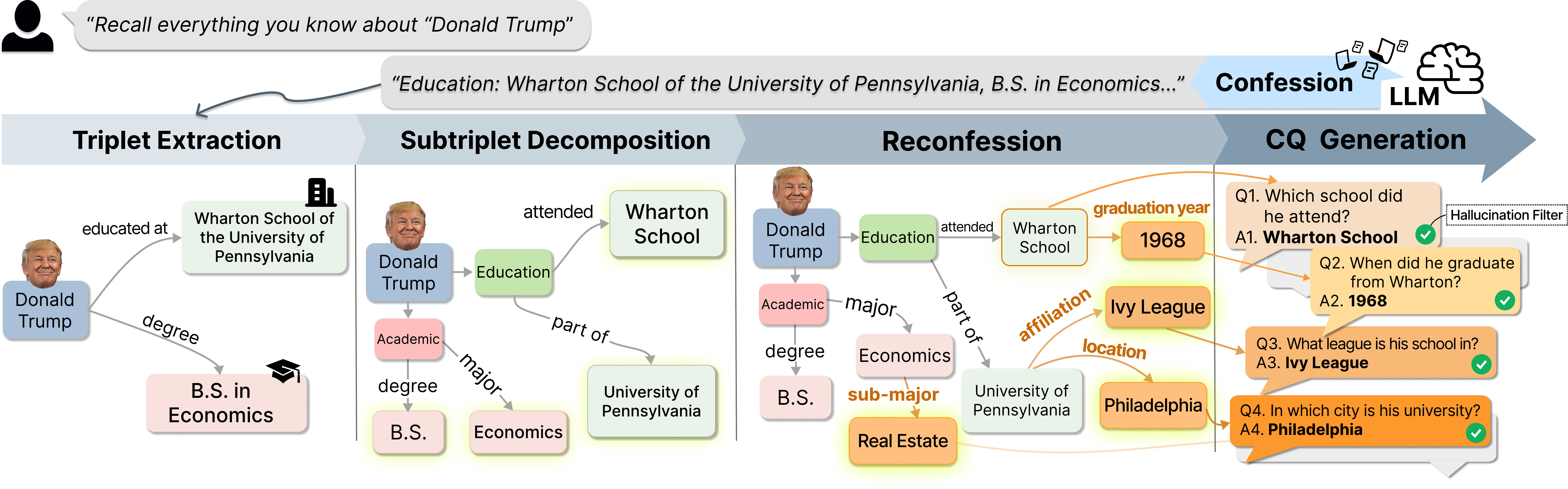}
  \caption{\textbf{Example of the CONFS pipeline applied to a confessed claim.}
  Using Donald Trump as an example entity, CONFS processes a single educational claim
  (\emph{Education: Wharton School of the University of Pennsylvania, B.S.\ in Economics}):
  the claim is transformed via triplet extraction and subtriplet decomposition,
  expanded through \emph{reconfession} to reveal attribute-level details,
  and finally converted into model-aligned competency questions.
  }
  \label{fig:framework}
\end{figure*}

\begin{table*}[t]
\centering
\small
\setlength{\tabcolsep}{2pt}
\renewcommand{\arraystretch}{1.25}
\setlength{\aboverulesep}{1.5pt}\setlength{\belowrulesep}{1.5pt}

\begin{adjustbox}{max width=\textwidth}
\begin{tabular}{l ccc ccc ccc ccc}
\toprule
& \multicolumn{3}{c}{\textbf{Forget set (10\%)}}
& \multicolumn{3}{c}{\textbf{Retain set (10\%) ($\uparrow$)}}
& \multicolumn{3}{c}{\textbf{Real Authors ($\uparrow$)}}
& \multicolumn{3}{c}{\textbf{World Facts ($\uparrow$)}} \\
\cmidrule(lr){2-4}\cmidrule(lr){5-7}\cmidrule(lr){8-10}\cmidrule(lr){11-13}
\textbf{Setting}
& Prob. ($\downarrow$) & R-L ($\downarrow$) & TR ($\uparrow$)
& Prob. & R-L & TR
& Prob. & R-L & TR
& Prob. & R-L & TR \\
\midrule

\textbf{Pre-trained}
& 0.990 & 0.979 & 0.513
& 0.989 & 0.981 & 0.470
& 0.061 & 0.943 & 0.580
& 0.017 & 0.875 & 0.559 \\

\midrule
\multicolumn{13}{c}{\textbf{GA}} \\
\midrule
Gold-standard
& 0.303\sd{.019} & 0.542\sd{.031} & 0.592\sd{.012}
& 0.331\sd{.021} & 0.566\sd{.042} & 0.430\sd{.011}
& 0.044\sd{.000} & 0.909\sd{.004} & 0.603\sd{.017}
& 0.032\sd{.005} & 0.907\sd{.004} & 0.582\sd{.001} \\
\specialrule{0.05em}{0em}{0em}

FreeRecall-QA
& 0.540\sd{.026} & 0.837\sd{.037} & \textbf{0.477}\sd{.002}
& 0.527\sd{.013} & \underline{0.784}\sd{.014} & 0.505\sd{.006}
& 0.006\sd{.003} & 0.730\sd{.028} & 0.531\sd{.003}
& \underline{0.005}\sd{.004} & 0.850\sd{.008} & 0.523\sd{.006} \\
RWKU-style
& 0.385\sd{.026} & 0.592\sd{.124} & 0.429\sd{.005}
& 0.368\sd{.019} & 0.380\sd{.195} & \textbf{0.537}\sd{.008}
& 0.011\sd{.003} & 0.820\sd{.087} & \underline{0.561}\sd{.004}
& 0.002\sd{.000} & 0.845\sd{.009} & 0.565\sd{.003} \\
CONFS \textit{w/o} Recon.
& 0.409\sd{.007} & 0.483\sd{.004} & 0.462\sd{.005}
& 0.517\sd{.009} & 0.562\sd{.007} & \underline{0.510}\sd{.002}
& 0.013\sd{.005} & 0.864\sd{.016} & 0.538\sd{.021}
& 0.004\sd{.000} & 0.856\sd{.019} & \underline{0.577}\sd{.008} \\
CONFS \textit{w/o} Halluc.
& \textbf{0.364}\sd{.057} & \underline{0.448}\sd{.017} & 0.464\sd{.001}
& \underline{0.589}\sd{.018} & 0.728\sd{.115} & 0.505\sd{.004}
& \underline{0.014}\sd{.010} & \underline{0.883}\sd{.044} & 0.553\sd{.018}
& 0.004\sd{.000} & \underline{0.867}\sd{.026} & 0.558\sd{.007} \\
\rowcolor{gray!12}
\textbf{CONFS}
& \underline{0.372}\sd{.023} & \textbf{0.374}\sd{.036} & \underline{0.467}\sd{.004}
& \textbf{0.621}\sd{.006} & \textbf{0.847}\sd{.005} & 0.509\sd{.000}
& \textbf{0.038}\sd{.009} & \textbf{0.891}\sd{.020} & \textbf{0.623}\sd{.012}
& \textbf{0.014}\sd{.002} & \textbf{0.876}\sd{.001} & \textbf{0.581}\sd{.015} \\

\midrule
\multicolumn{13}{c}{\textbf{GD}} \\
\midrule
Gold-standard
& 0.448\sd{.021} & 0.518\sd{.014} & 0.504\sd{.003}
& 0.843\sd{.024} & 0.655\sd{.014} & 0.480\sd{.012}
& 0.018\sd{.004} & 0.890\sd{.003} & 0.497\sd{.023}
& 0.004\sd{.003} & 0.854\sd{.002} & 0.505\sd{.004} \\
\specialrule{0.05em}{0em}{0em}

FreeRecall-QA
& 0.716\sd{.027} & 0.860\sd{.036} & \underline{0.496}\sd{.002}
& 0.759\sd{.028} & \textbf{0.882}\sd{.008} & 0.480\sd{.013}
& \underline{0.039}\sd{.012} & \textbf{0.893}\sd{.014} & 0.582\sd{.038}
& 0.006\sd{.002} & 0.870\sd{.004} & 0.566\sd{.015} \\

RWKU-style
& 0.633\sd{.004} & 0.578\sd{.146} & 0.435\sd{.023}
& 0.768\sd{.106} & 0.599\sd{.153} & \textbf{0.539}\sd{.001}
& 0.017\sd{.003} & 0.840\sd{.014} & 0.519\sd{.047}
& 0.005\sd{.002} & 0.870\sd{.005} & 0.593\sd{.043} \\

CONFS \textit{w/o} Recon.
& 0.604\sd{.012} & 0.575\sd{.014} & 0.491\sd{.004}
& \underline{0.860}\sd{.013} & 0.810\sd{.014} & 0.483\sd{.023}
& \underline{0.039}\sd{.018} & 0.882\sd{.027} & 0.574\sd{.024}
& 0.018\sd{.006} & \textbf{0.881}\sd{.014} & 0.589\sd{.032} \\

CONFS \textit{w/o} Halluc.
& \underline{0.556}\sd{.014} & \underline{0.568}\sd{.027} & 0.491\sd{.013}
& \textbf{0.892}\sd{.012} & 0.799\sd{.014} & \underline{0.495}\sd{.011}
& 0.038\sd{.017} & \underline{0.886}\sd{.026} & \underline{0.587}\sd{.037}
& \underline{0.019}\sd{.006} & \underline{0.872}\sd{.013} & \underline{0.608}\sd{.039} \\

\rowcolor{gray!12}
\textbf{CONFS}
& \textbf{0.521}\sd{.007} & \textbf{0.497}\sd{.007} & \textbf{0.504}\sd{.001}
& 0.781\sd{.003} & \underline{0.833}\sd{.052} & 0.490\sd{.001}
& \textbf{0.062}\sd{.002} & 0.884\sd{.034} & \textbf{0.600}\sd{.002}
& \textbf{0.020}\sd{.001} & 0.825\sd{.000} & \textbf{0.623}\sd{.004} \\

\midrule
\multicolumn{13}{c}{\textbf{NPO}} \\
\midrule
Gold-standard
& 0.318\sd{.015} & 0.519\sd{.002} & 0.613\sd{.019}
& 0.353\sd{.041} & 0.540\sd{.024} & 0.412\sd{.001}
& 0.042\sd{.008} & 0.901\sd{.005} & 0.631\sd{.009}
& 0.044\sd{.004} & 0.908\sd{.001} & 0.595\sd{.006} \\
\specialrule{0.05em}{0em}{0em}

FreeRecall-QA
& 0.562\sd{.018} & 0.632\sd{.003} & 0.479\sd{.006}
& 0.454\sd{.015} & 0.460\sd{.028} & \underline{0.504}\sd{.008}
& 0.015\sd{.002} & 0.899\sd{.015} & 0.566\sd{.003}
& 0.004\sd{.000} & \underline{0.879}\sd{.005} & 0.547\sd{.007} \\
RWKU-style
& 0.634\sd{.013} & 0.606\sd{.056} & \underline{0.507}\sd{.021}
& 0.580\sd{.004} & \underline{0.612}\sd{.050} & 0.488\sd{.012}
& \textbf{0.034}\sd{.002} & \textbf{0.917}\sd{.004} & \underline{0.584}\sd{.003}
& {0.011}\sd{.003} & 0.877\sd{.012} & 0.561\sd{.005} \\

CONFS \textit{w/o} Recon.
& 0.534\sd{.063} & 0.565\sd{.014} & 0.476\sd{.024}
& \underline{0.586}\sd{.018} & 0.593\sd{.004} & 0.493\sd{.022}
& 0.030\sd{.012} & 0.875\sd{.005} & 0.551\sd{.042}
& \underline{0.031}\sd{.018} & 0.839\sd{.013} & \underline{0.589}\sd{.037} \\

CONFS \textit{w/o} Halluc.
& \textbf{0.359}\sd{.027} & \underline{0.503}\sd{.012} & \textbf{0.520}\sd{.072}
& 0.499\sd{.018} & 0.563\sd{.013} & 0.503\sd{.004}
& 0.030\sd{.017} & 0.860\sd{.036} & \textbf{0.585}\sd{.038}
& 0.027\sd{.011} & 0.856\sd{.019} & \textbf{0.592}\sd{.014} \\
\rowcolor{gray!12}
\textbf{CONFS}
& \underline{0.443}\sd{.030} & \textbf{0.488}\sd{.042} & 0.473\sd{.000}
& \textbf{0.681}\sd{.003} & \textbf{0.620}\sd{.003} & \textbf{0.510}\sd{.000}
& \underline{0.031}\sd{.003} & \underline{0.911}\sd{.003} & 0.527\sd{.004}
& \textbf{0.034}\sd{.001} & \textbf{0.882}\sd{.004} & 0.566\sd{.004} \\

\midrule
\multicolumn{13}{c}{\textbf{RT}} \\
\midrule
Gold-standard
& 0.665\sd{.028} & 0.020\sd{.018} & 0.559\sd{.037}
& 0.684\sd{.021} & 0.021\sd{.012} & 0.425\sd{.026}
& 0.082\sd{.046} & 0.019\sd{.017} & 0.513\sd{.038}
& 0.053\sd{.027} & 0.017\sd{.016} & 0.539\sd{.032} \\
\specialrule{0.05em}{0em}{0em}

FreeRecall-QA
& 0.769\sd{.013} & 0.032\sd{.002} & 0.550\sd{.001}
& \textbf{0.781}\sd{.019} & 0.040\sd{.007} & 0.435\sd{.000}
& 0.036\sd{.000} & 0.042\sd{.002} & 0.439\sd{.006}
& 0.017\sd{.001} & 0.017\sd{.002} & 0.519\sd{.003} \\

RWKU-style
& 0.723\sd{.018} & 0.034\sd{.012} & 0.552\sd{.027}
& 0.713\sd{.026} & 0.028\sd{.017} & 0.427\sd{.019}
& 0.047\sd{.016} & 0.033\sd{.026} & 0.501\sd{.036}
& \underline{0.032}\sd{.011} & 0.028\sd{.018} & 0.522\sd{.028} \\

CONFS \textit{w/o} Recon.
& 0.725\sd{.026} & \underline{0.027}\sd{.018} & 0.555\sd{.021}
& 0.732\sd{.019} & \underline{0.041}\sd{.027} & 0.430\sd{.026}
& \underline{0.060}\sd{.037} & \underline{0.043}\sd{.018} & \underline{0.505}\sd{.028}
& \textbf{0.037}\sd{.017} & 0.035\sd{.026} & \underline{0.531}\sd{.036} \\

CONFS \textit{w/o} Halluc.
& \underline{0.676}\sd{.019} & \textbf{0.019}\sd{.011} & \textbf{0.566}\sd{.018}
& 0.730\sd{.026} & \textbf{0.049}\sd{.017} & \underline{0.438}\sd{.027}
& 0.056\sd{.018} & 0.032\sd{.036} & 0.485\sd{.028}
& 0.027\sd{.017} & \underline{0.061}\sd{.027} & 0.515\sd{.019} \\

\rowcolor{gray!12}
\textbf{CONFS}
& \textbf{0.670}\sd{.001} & 0.030\sd{.001} & \underline{0.565}\sd{.000}
& \underline{0.777}\sd{.001} & 0.023\sd{.004} & \textbf{0.449}\sd{.001}
& \textbf{0.066}\sd{.000} & \textbf{0.050}\sd{.003} & \textbf{0.533}\sd{.002}
& 0.028\sd{.000} & \textbf{0.065}\sd{.004} & \textbf{0.533}\sd{.004} \\

\bottomrule
\end{tabular}
\end{adjustbox}

\caption{\textbf{TOFU (10\%) unlearning results under different forget-set constructions},
reported as mean$\pm$std over 3 seeds.
The best and second-best data-blind settings are highlighted in \textbf{bold} and \underline{underline}, respectively.}
\label{tab:tofu_unlearning_results}
\end{table*}

\section{Method}
\label{sec:method}

We present the \textbf{CONfession-to-Forget-Set (CONFS)} framework, which addresses \emph{forget-set misalignment} by identifying a model’s memorized knowledge about a target entity $E$ and constructing a model-aligned forget set.
We consider a \emph{data-blind} setting, in which the pre-training data are inaccessible and the information to be removed from the model is not specified in advance.

\paragraph{Operational Definition of Memorized Knowledge.}
We call a fact \emph{memorized} if the target model reproduces it under elicitation and does so consistently across stochastic samples.
CONFS operationalizes this criterion through elicitation stages (\emph{confession} and \emph{reconfession}), followed by a consistency check (\emph{hallucination verification}).
This criterion is behavioral and closely related to extractable memorization~\citep{carlini2021extracting}, but distinct from training-data provenance: it does not establish which document taught the fact, and a stable but incorrect output remains in scope because the model will still disclose it to a user.
We adopt this behavioral criterion for two reasons. First, unlearning edits parameters to change what a model discloses, not which document it read. Second, provenance is unavailable in a data-blind setting, where recovering it post hoc reduces to membership inference, which performs near chance on LLM pre-training data~\citep{duan2024membership}.

A key challenge is that such knowledge is latent and distributed, making its exact scope difficult to inspect directly.
Even when elicited through prompting, it appears as unstructured claims and may expose only partial information.

\paragraph{Structural Formalization via Triplets.} 
To address the lack of explicit structure in elicited claims, we formalize them into a discrete set of Subject-Relation-Object (SRO) triplets.
Unlike \citet{jin2024rwku}, which treats unlearning targets as raw text, we discretize ambiguous semantic content into verifiable, atomic units for fine-grained unlearning.

\paragraph{Exhaustive Probing via Reconfession.} 
A single query rarely reveals everything a model knows about a target entity. We therefore add a \emph{reconfession} step after the initial \emph{confession} to recover this missed knowledge.
As illustrated in Fig.~\ref{fig:framework}, core attributes from the initial \emph{confession} serve as cues for recursively probing hidden details, resulting in a more comprehensive and faithful forget set.

\subsection{Confession: Exposing Model Memory}

The \emph{confession} stage exposes the knowledge the model has memorized about a target entity $E$.
By prompting the model with only the entity name and no external information, we obtain a set of raw natural-language claims $\mathcal{C}_E = \{c_1, \dots, c_n\}$ that reflect the scope of this memorized knowledge.

\begin{table*}[t]
\centering
\small
\setlength{\tabcolsep}{2pt}
\renewcommand{\arraystretch}{1.25}
\setlength{\aboverulesep}{1.5pt}\setlength{\belowrulesep}{1.5pt}

\begin{adjustbox}{max width=\textwidth}
\begin{tabular}{l ccc ccc ccc ccc}
\toprule
& \multicolumn{3}{c}{\textbf{Forget set (10\%)}} 
& \multicolumn{3}{c}{\textbf{Retain set (10\%) ($\uparrow$)}} 
& \multicolumn{3}{c}{\textbf{Real Faces ($\uparrow$)}} 
& \multicolumn{3}{c}{\textbf{Real World ($\uparrow$)}} \\
\cmidrule(lr){2-4}\cmidrule(lr){5-7}\cmidrule(lr){8-10}\cmidrule(lr){11-13}
\textbf{Setting}
& Prob. ($\downarrow$) & R-L ($\downarrow$) & TR ($\uparrow$)
& Prob. & R-L & TR
& Prob. & R-L & TR
& Prob. & R-L & TR \\
\midrule

\textbf{Pre-trained}
& 0.183 & 0.402 & 0.740
& 0.181 & 0.330 & 0.138
& 0.254 & 0.142 & 0.106
& 0.335 & 0.517 & 0.124 \\

\midrule
\multicolumn{13}{c}{\textbf{GA}} \\
\midrule

Gold-standard
& 0.007\sd{.009} & 0.300\sd{.007} & 0.569\sd{.012}
& 0.014\sd{.019} & 0.303\sd{.058} & 0.003\sd{.004}
& 0.003\sd{.002} & 0.035\sd{.022} & 0.124\sd{.047}
& 0.028\sd{.018} & 0.265\sd{.027} & 0.230\sd{.009} \\
\specialrule{0.05em}{0em}{0em}

FreeRecall-QA
& \textbf{0.004}\sd{.005} & \underline{0.355}\sd{.029} & 0.393\sd{.032}
& 0.004\sd{.005} & 0.293\sd{.026} & 0.206\sd{.021}
& 0.003\sd{.003} & \underline{0.037}\sd{.032} & 0.109\sd{.014}
& 0.006\sd{.004} & 0.246\sd{.008} & 0.260\sd{.041} \\

CONFS \textit{w/o} Recon.
& 0.015\sd{.007} & 0.396\sd{.071} & 0.429\sd{.017}
& 0.018\sd{.009} & \underline{0.388}\sd{.046} & \underline{0.253}\sd{.011}
& 0.005\sd{.005} & \textbf{0.040}\sd{.043} & 0.179\sd{.023}
& 0.010\sd{.013} & 0.323\sd{.067} & \textbf{0.357}\sd{.006} \\

CONFS \textit{w/o} Halluc.
& 0.016\sd{.022} & 0.395\sd{.054} & \underline{0.444}\sd{.028}
& \underline{0.023}\sd{.010} & 0.383\sd{.039} & \textbf{0.254}\sd{.057}
& \underline{0.006}\sd{.007} & 0.031\sd{.043} & \textbf{0.219}\sd{.073}
& \underline{0.033}\sd{.031} & \underline{0.324}\sd{.019} & \underline{0.354}\sd{.042} \\

\rowcolor{gray!12}
\textbf{CONFS}
& \underline{0.014}\sd{.019} & \textbf{0.322}\sd{.022} & \textbf{0.577}\sd{.048}
& \textbf{0.031}\sd{.036} & \textbf{0.398}\sd{.051} & 0.233\sd{.072}
& \textbf{0.007}\sd{.009} & 0.036\sd{.009} & \underline{0.196}\sd{.066}
& \textbf{0.074}\sd{.074} & \textbf{0.383}\sd{.027} & 0.240\sd{.070} \\

\midrule
\multicolumn{13}{c}{\textbf{GD}} \\
\midrule

Gold-standard
& 0.183\sd{.014} & 0.329\sd{.068} & 0.284\sd{.071}
& 0.168\sd{.016} & 0.263\sd{.047} & 0.373\sd{.033}
& 0.257\sd{.053} & 0.140\sd{.044} & 0.238\sd{.008}
& 0.332\sd{.063} & 0.249\sd{.024} & 0.168\sd{.049} \\
\specialrule{0.05em}{0em}{0em}

FreeRecall-QA
& 0.144\sd{.052} & 0.242\sd{.059} & 0.066\sd{.029}
& 0.168\sd{.011} & \underline{0.231}\sd{.062} & 0.598\sd{.020}
& 0.254\sd{.040} & 0.054\sd{.013} & \underline{0.297}\sd{.056}
& \underline{0.336}\sd{.064} & 0.242\sd{.046} & 0.346\sd{.006} \\

CONFS \textit{w/o} Recon.
& 0.152\sd{.032} & 0.234\sd{.026} & 0.062\sd{.054}
& 0.166\sd{.018} & 0.184\sd{.021} & 0.568\sd{.034}
& \underline{0.256}\sd{.050} & \underline{0.072}\sd{.030} & 0.223\sd{.010}
& 0.335\sd{.017} & \underline{0.262}\sd{.007} & 0.378\sd{.012} \\

CONFS \textit{w/o} Halluc.
& \underline{0.131}\sd{.048} & \textbf{0.193}\sd{.038} & \underline{0.099}\sd{.028}
& \underline{0.178}\sd{.041} & \textbf{0.259}\sd{.037} & \underline{0.599}\sd{.067}
& 0.251\sd{.073} & 0.065\sd{.031} & 0.296\sd{.027}
& \textbf{0.342}\sd{.060} & 0.256\sd{.023} & \underline{0.423}\sd{.019} \\

\rowcolor{gray!12}
\textbf{CONFS}
& \textbf{0.114}\sd{.051} & \underline{0.218}\sd{.069} & \textbf{0.115}\sd{.043}
& \textbf{0.184}\sd{.072} & 0.206\sd{.047} & \textbf{0.622}\sd{.053}
& \textbf{0.257}\sd{.039} & \textbf{0.079}\sd{.049} & \textbf{0.308}\sd{.058}
& 0.331\sd{.052} & \textbf{0.274}\sd{.009} & \textbf{0.487}\sd{.033} \\

\midrule
\multicolumn{13}{c}{\textbf{NPO}} \\
\midrule

Gold-standard
& 0.157\sd{.066} & 0.373\sd{.022} & 0.567\sd{.046}
& 0.158\sd{.061} & 0.375\sd{.016} & 0.207\sd{.054}
& 0.112\sd{.057} & 0.044\sd{.024} & 0.159\sd{.063}
& 0.278\sd{.050} & 0.346\sd{.014} & 0.245\sd{.048} \\
\specialrule{0.05em}{0em}{0em}

FreeRecall-QA
& 0.188\sd{.042} & 0.375\sd{.074} & \underline{0.551}\sd{.059}
& 0.182\sd{.029} & \underline{0.373}\sd{.062} & 0.199\sd{.056}
& 0.234\sd{.008} & 0.044\sd{.051} & \underline{0.162}\sd{.064}
& 0.330\sd{.070} & 0.328\sd{.060} & \underline{0.273}\sd{.047} \\

CONFS \textit{w/o} Recon.
& 0.191\sd{.068} & 0.361\sd{.053} & 0.522\sd{.032}
& 0.180\sd{.020} & 0.320\sd{.036} & \underline{0.202}\sd{.026}
& \underline{0.239}\sd{.011} & \underline{0.063}\sd{.071} & 0.152\sd{.044}
& \underline{0.335}\sd{.018} & 0.333\sd{.058} & 0.267\sd{.040} \\

CONFS \textit{w/o} Halluc.
& \textbf{0.181}\sd{.067} & \underline{0.357}\sd{.073} & 0.546\sd{.013}
& \textbf{0.188}\sd{.069} & 0.350\sd{.021} & \textbf{0.206}\sd{.017}
& 0.228\sd{.023} & 0.032\sd{.006} & \textbf{0.168}\sd{.061}
& 0.333\sd{.038} & \underline{0.341}\sd{.041} & \textbf{0.278}\sd{.049} \\

\rowcolor{gray!12}
\textbf{CONFS}
& \underline{0.186}\sd{.010} & \textbf{0.340}\sd{.052} & \textbf{0.660}\sd{.046}
& \underline{0.187}\sd{.057} & \textbf{0.445}\sd{.019} & 0.162\sd{.007}
& \textbf{0.250}\sd{.072} & \textbf{0.103}\sd{.066} & 0.125\sd{.034}
& \textbf{0.340}\sd{.012} & \textbf{0.382}\sd{.054} & 0.213\sd{.074} \\

\midrule
\multicolumn{13}{c}{\textbf{RT}} \\
\midrule

Gold-standard
& 0.175\sd{.022} & 0.009\sd{.004} & 0.960\sd{.009}
& 0.177\sd{.030} & 0.000\sd{.000} & 0.027\sd{.014}
& 0.245\sd{.037} & 0.001\sd{.001} & 0.008\sd{.004}
& 0.338\sd{.028} & 0.022\sd{.008} & 0.012\sd{.002} \\
\specialrule{0.05em}{0em}{0em}

FreeRecall-QA
& \underline{0.176}\sd{.011} & 0.006\sd{.003} & 0.952\sd{.016}
& \textbf{0.182}\sd{.013} & 0.004\sd{.003} & \textbf{0.034}\sd{.006}
& 0.247\sd{.024} & \textbf{0.000}\sd{.000} & 0.008\sd{.004}
& 0.330\sd{.043} & 0.010\sd{.010} & 0.018\sd{.007} \\

CONFS \textit{w/o} Recon.
& 0.177\sd{.020} & \textbf{0.001}\sd{.001} & 0.961\sd{.031}
& 0.174\sd{.027} & \underline{0.008}\sd{.010} & 0.029\sd{.009}
& \textbf{0.255}\sd{.050} & \textbf{0.000}\sd{.000} & \textbf{0.019}\sd{.014}
& \underline{0.336}\sd{.018} & 0.012\sd{.008} & 0.015\sd{.004} \\

CONFS \textit{w/o} Halluc.
& 0.179\sd{.033} & \underline{0.002}\sd{.002} & \underline{0.962}\sd{.011}
& \underline{0.175}\sd{.039} & 0.003\sd{.001} & 0.025\sd{.021}
& \underline{0.250}\sd{.017} & \textbf{0.000}\sd{.000} & \underline{0.013}\sd{.013}
& 0.331\sd{.023} & \underline{0.028}\sd{.006} & \underline{0.021}\sd{.010} \\

\rowcolor{gray!12}
\textbf{CONFS}
& \textbf{0.165}\sd{.007} & 0.003\sd{.003} & \textbf{0.966}\sd{.019}
& 0.170\sd{.012} & \textbf{0.040}\sd{.022} & \underline{0.031}\sd{.009}
& 0.246\sd{.042} & \textbf{0.000}\sd{.000} & 0.012\sd{.002}
& \textbf{0.338}\sd{.029} & \textbf{0.118}\sd{.063} & \textbf{0.026}\sd{.014} \\

\bottomrule
\end{tabular}
\end{adjustbox}

\caption{\textbf{CLEAR (10\%) unlearning results under different forget-set constructions},
reported as mean$\pm$std over 3 seeds.
The best and second-best data-blind settings are highlighted in \textbf{bold} and \underline{underline}, respectively.
RWKU-style is omitted for the multimodal setting.}
\label{tab:clear_unlearning_results}
\end{table*}

\subsection{Triplet Extraction and Subtriplet Decomposition}
After the \emph{confession} stage, the model's memorized knowledge appears as unstructured natural-language claims that may encode multiple factual units.
We discretize them into verifiable, atomic units, so that each forget-set element corresponds to a single, well-defined fact to be erased.

\paragraph{Triplet Extraction.}
We convert each claim $c \in \mathcal{C}_E$ into an SRO triplet $(s, r, o)$.
The Subject $s$ is fixed as the target entity $E$.
The Relation $r$ is defined as a noun-based attribute characterizing the entity, rather than a surface-level verb.
The Object $o$ corresponds to the explicit entity-specific value stated in the claim.

\paragraph{Subtriplet Decomposition.}
Since a single triplet may still encode multiple factual attributes within a composite Object, we decompose non-atomic triplets into fine-grained subtriplets, retaining a triplet unchanged when no further decomposition is possible.
All resulting triplets and subtriplets derived from a claim are treated as \emph{preliminary leaf candidates}, each taking the form $(E, r, o)$.

\subsection{Reconfession: Attribute-Level Knowledge Probing}
To ensure exhaustive unlearning without uncontrolled expansion, we introduce \emph{reconfession}.
Due to sampling bias, a single query often fails to expose all attribute-level knowledge of an entity,
even after extracting initial leaf candidates $(E,r,o)$.
\emph{Reconfession} addresses this limitation by selectively probing additional attributes, thereby expanding the coverage of the model's memorized knowledge.

\paragraph{Reconfession Decision Criterion.}
\emph{Reconfession} applies when the Object $o$ of a leaf triplet $(E, r, o)$ functions as a \textit{sub-entity} under Relation $r$.
In this case, additional attributes $p$ of $o$ may exist but remain unrepresented.
For example, $(E, \textit{authored}, o)$ identifies a book whose publication year is also memorized, yielding an expanded leaf $(E,\textit{authored},o,\textit{publication\_year},v)$. Such triplets are marked as reconfessable.

\paragraph{Selective Attribute Probing.}
For each reconfessable leaf triplet $(E, r, o)$ with an identified attribute $p$, we query the model for the corresponding value $v$ using only the existing leaf information.
The model either returns a concrete value $v$, treated as exposed memorized knowledge, or responds with \texttt{UNKNOWN}, which terminates further expansion.
After \emph{reconfession}, leaf representations take one of two forms: base leaves $(E,r,o)$ and attribute leaves $(E,r,o,p,v)$.

\subsection{Competency Question Generation}
In the final stage, finalized leaf representations are converted into \emph{competency questions} (CQs), which serve as direct inputs for unlearning.
Each CQ targets exactly one leaf-level knowledge unit and is constructed only from the information explicitly contained in that representation, without introducing external knowledge.

\paragraph{Generation Rules.}
For each base leaf $(E, r, o)$, we generate one question with $o$ as the answer.
For each attribute-expanded leaf $(E, r, o, p, v)$, we generate two questions, targeting $o$ and $v$, respectively.
The final competency question set aggregates all questions generated across the claims $c \in \mathcal{C}_E$.
We use GPT-4o for the above triplet structuring and competency question generation, without external knowledge (Appendix~\ref{CONFS_prompt}).

\paragraph{Hallucination Verification.}
Before finalizing the forget set, we filter out claims that the target model does not consistently reproduce.
Following SelfCheckGPT~\citep{manakul2023selfcheckgpt}, we sample multiple stochastic answers from the target model for each CQ. We retain a QA pair only when its average contradiction probability, scored by a DeBERTa-v3-large NLI model~\citep{he2021debertav3}, falls below $\tau$.
We set $\tau{=}0.7$ and provide robustness analysis in Appendix~\ref{app:forgetset_quality}.
The complete CONFS algorithm is summarized in Appendix~\ref{alg:confs}.

\begin{table*}[t]
\centering
\small
\setlength{\tabcolsep}{10pt}
\begin{tabular}{l ccc cc}
\toprule
& \multicolumn{3}{c}{\textbf{Out-of-Knowledge}} & \multicolumn{2}{c}{\textbf{Under}} \\
\cmidrule(lr){2-4}\cmidrule(lr){5-6}
\textbf{Forget set}
 & $\mathcal{D}_{N+}$ share $\downarrow$
 & $\cos(g_{N+},g_{retain})$ $\downarrow$
 & $\Delta\mathcal{L}(\mathcal{D}_{retain})$ $\downarrow$
 & $\widehat{E}_{\mathrm{tf}}$ $\uparrow$
 & $\widehat{E}_{\mathrm{eg}}\!\to\!0$ \\
\midrule
FreeRecall-QA & 91\% & 0.84 & $+3.53$ & 0.57 & 0.03 \\
RWKU-style & 68\% & 0.46 & $+0.31$ & 0.97 & $-0.01$ \\
\rowcolor{gray!15}
CONFS & \textbf{54\%} & \textbf{0.28} & $\mathbf{+0.11}$ & \textbf{0.97} & $\mathbf{-0.00}$ \\
\bottomrule
\end{tabular}
\caption{\textbf{Misalignment diagnostics on the main TOFU forget sets.}
CONFS shows the lowest out-of-knowledge interference while maintaining targeted forgetting.}
\label{tab:misalign_main}
\end{table*}

\section{Experiments}

\subsection{Experiment Setups}\label{sec:exp_setups}

\textbf{Benchmarks.}
We evaluate our method on synthetic, multimodal, and real-entity benchmarks: TOFU~\cite{maini2024tofu}, CLEAR~\cite{dontsov2025clear}, and RWKU~\cite{jin2024rwku}.
\textbf{TOFU} is a synthetic benchmark with 200 fictitious author profiles and LLM-generated QA examples.
Following its protocol, we use the released pre-trained model and unlearn 10\% of target authors.
\textbf{CLEAR} extends TOFU to the vision-language setting with textual QA and synthetic face images. We use the same 10\% target-persona protocol.
\textbf{RWKU} targets real-world knowledge removal for public figures, drawn from 200 candidates by Wikipedia page-view popularity.
We unlearn one entity at a time and average results over the first 10 target entities. We use LLaMA-2-7B-Chat~\cite{touvron2023llama} as the base model for TOFU and RWKU, and LLaVA-1.5-7B~\cite{liu2024improved} for CLEAR.

\paragraph{Forget set constructions and comparisons.}
We evaluate how closely different \emph{data-blind} forget-set constructions approximate the \emph{Gold-standard}, where unlearning uses the original benchmark-provided target data corresponding to the injected knowledge.
We compare:
(i) \textbf{FreeRecall-QA}: unstructured free recall of factual QA pairs by a pre-trained LLM (Appendix~\ref{nameQA});
(ii) \textbf{RWKU-style}: forget sets constructed using the RWKU probing pipeline~\cite{jin2024rwku};
(iii) \textbf{CONFS \textit{w/o} Recon.}: CONFS without the \emph{reconfession} step;
(iv) \textbf{CONFS \textit{w/o} Halluc.}: CONFS with \emph{reconfession} but without hallucination verification.

\paragraph{Unlearning objectives and metrics.}
Given a forget set, we update model parameters using four representative loss-based unlearning objectives:
Gradient Ascent (GA; \citeauthor{jang2023knowledge}, \citeyear{jang2023knowledge}),
Gradient Difference (GD; \citeauthor{liu2022continual}, \citeyear{liu2022continual}),
Negative Preference Optimization (NPO; \citeauthor{zhang2024negative}, \citeyear{zhang2024negative}),
and Rejection Tuning (RT; IDK-style; \citeauthor{maini2024tofu}, \citeyear{maini2024tofu}).
We follow the official evaluation protocols of TOFU, CLEAR, and RWKU, with implementation details and metric definitions in Appendix~\ref{imp}.

\subsection{Experiment Results}
\textbf{TOFU Benchmark.}
Table~\ref{tab:tofu_unlearning_results} reports results across GA, GD, NPO, and RT with different forget-set constructions.
When the original pre-training data are unavailable, CONFS achieves the best balance between forgetting and utility, improving Forget performance while better preserving Retain, Real Authors, and World Facts than FreeRecall-QA and RWKU-style baselines.
Additional forget-set quality analyses and qualitative examples are provided in Appendices~\ref{app:forgetset_quality} and~\ref{app:forgetset_quantity}.
{Across the pipeline stages, \emph{reconfession} raises Recall and forget-set size, and \emph{hallucination verification} then raises Precision and F1 while shrinking the set (Table~\ref{tab:forgetset_quality}); replacing the GPT-4o structurer with Qwen2.5-7B-Instruct or Llama-3.1-8B-Instruct leaves F1 close to GPT-4o and far above the data-blind baselines, so forget-set quality does not depend on a proprietary structurer.}

\paragraph{Misalignment Diagnostics on the Main Forget Sets.}
Table~\ref{tab:misalign_main} applies the Section~\ref{3} diagnostics to the same three constructions.
For each one we contrast the requested forget set with the Gold TOFU forget set, consisting of the QA pairs injected during pre-training, and partition the data as in Section~\ref{3}.
The diagnostics are computed on the TOFU-finetuned model, and the partition follows the same fact-matching judgments as Table~\ref{tab:forgetset_quality}.
The $\mathcal{D}_{N+}$ share is the fraction of requested facts outside the Gold set, and $\Delta\mathcal{L}(\mathcal{D}_{retain})$ is the measured retain-loss increase over the pre-unlearning baseline of $0.45$.
The two Under-setting effects, which use $\mathbf{F}=\mathcal{D}_{L+}$, are normalized by $\eta\|g_{L+}\|^{2}$, i.e., $\widehat{E}_{\mathrm{tf}}:=E_{\mathrm{tf}}/\eta\|g_{L+}\|^{2}$ and likewise for $\widehat{E}_{\mathrm{eg}}$, so that forget sets of different sizes are comparable.
Out-of-Knowledge collateral damage scales monotonically with the $\mathcal{D}_{N+}$ share, and the directly measured retain damage follows the same order, matching the forget-set F1 ranking of Table~\ref{tab:forgetset_quality}.
In the Under setting, targeted forgetting is near-ideal for CONFS but markedly weaker for FreeRecall-QA, while $\widehat{E}_{\mathrm{eg}}$ remains near zero across the three constructions, so forgetting does not spread to the memorized facts left out of the request.
Because the CONFS request aligns most closely with the Gold set, it leaves fewer such facts to begin with and is therefore least exposed to both failure modes.

\paragraph{CLEAR Benchmark.}
Table~\ref{tab:clear_unlearning_results} shows that CONFS maintains a favorable balance between forgetting and utility on CLEAR, demonstrating its robustness in multimodal LLM settings.

\paragraph{RWKU Benchmark.}
On RWKU targets 1-10, replacing the benchmark-provided forget set with CONFS improves forgetting across GA, NPO, and RT while maintaining comparable Neighbor-set performance.
Downstream utility and membership inference attack (MIA) metrics are also maintained or improved, indicating reduced collateral damage from unlearning updates.
Detailed results are reported in Appendix~\ref{app:RWKU}.

\section{Conclusion}
We show that \emph{forget-set misalignment}, the mismatch between the forget set and the model's memory, causes unlearning failures.
We identify two failure modes, \emph{Under Unlearning} and \emph{Out-of-Knowledge Unlearning}, through gradient analysis and empirical evaluation.
To address this, we propose \textbf{CONFS}, which constructs model-aligned forget sets for targeted forgetting that preserves utility. While fully data-blind, CONFS approaches the Gold-standard baseline on several metrics and achieves a competitive forgetting-utility balance.

\newpage
\section*{Limitations}

This work focuses on factual knowledge about entities, where memorized information can be structured into Subject-Relation-Object (SRO) units and converted into competency questions.
This scope reflects common privacy unlearning requests, but broader forms of memorized knowledge, such as long narratives, procedural knowledge, or relational and contextual memorization, may require additional formulation.
Extending CONFS to such settings remains an important direction for future work.

\section*{Ethical Considerations}
This work studies LLM unlearning to remove privacy-sensitive information
that a model has memorized about an individual. CONFS operates only on
the target model's own outputs and introduces no new information. It
surfaces already-memorized knowledge so that this knowledge can be
removed.
Our experiments use publicly available benchmarks. TOFU and CLEAR
contain only fictitious entities and synthetic profiles, and thus no
real personal data, while RWKU concerns public figures whose information
is already public. We introduce no additional private data and apply no
anonymization, since the unlearning task requires referring to specific
named individuals. We observed no offensive content in the benchmarks we
use.
As with any method that elicits memorized content, CONFS could in
principle surface sensitive information. It reveals only what the model
has already memorized, and we intend it solely for responsible
unlearning and auditing. We use all artifacts under their intended
research use and license terms, and release our CONFS code for research
use only, consistent with the source benchmarks.

\section*{Acknowledgments}
This research was supported by the National Research Foundation of Korea (NRF) grant funded by the Korea government (MSIT) (RS-2025-00556289), and by the "Advanced GPU Utilization Support Program" funded by the Government of the Republic of Korea (Ministry of Science and ICT).

\bibliography{custom}

\clearpage
\appendix

\section{CONFS Algorithm}
\label{alg:confs}

\begin{algorithm}
\caption{CONFS Forget-Set Construction}
\small
\begin{algorithmic}[1]
\Require target entity $E$; target model $M$; structuring model $G$ (GPT-4o);
         confession samples $K\!=\!5$; verification samples $N\!=\!5$;
         threshold $\tau\!=\!0.7$; NLI model $\Phi$
\Ensure model-aligned forget set $\mathcal{F}$

\State $\mathcal{C}\gets\emptyset$
\For{$k=1,\dots,K$}
  \State $\mathcal{C}\gets\mathcal{C}\cup\textsc{Confess}(M,E)$
  \Comment{sample raw claims}
\EndFor

\State $\mathcal{L}\gets\emptyset$ \Comment{leaf representations}

\ForAll{$c\in\mathcal{C}$}
  \State $\mathcal{T}_c\gets\textsc{ExtractTriplets}(G,c,E)$
  \ForAll{$t\in\mathcal{T}_c$}
    \State $\mathcal{L}\gets\mathcal{L}\cup\textsc{Decompose}(G,t)$
    \Comment{atomic base leaves}
  \EndFor
\EndFor

\State $\mathcal{L}_{\mathrm{base}}\gets\mathcal{L}$
\ForAll{$\ell=(E,r,o)\in\mathcal{L}_{\mathrm{base}}$}
  \State $p\gets\textsc{DecideReconfession}(G,\ell)$
  \Comment{attribute to probe, or NONE}
  \If{$p\neq\textsc{NONE}$}
    \State $v\gets\textsc{Reconfess}(M,E,r,o,p)$
    \If{$v\neq\textsc{UNKNOWN}$}
      \State $\mathcal{L}\gets\mathcal{L}\cup\{(E,r,o,p,v)\}$
      \Comment{attribute-expanded leaf}
    \EndIf
  \EndIf
\EndFor

\State $\mathcal{Q}\gets\emptyset$
\ForAll{$\ell\in\mathcal{L}$}
  \State $\mathcal{Q}\gets\mathcal{Q}\cup\textsc{GenerateCQs}(G,\ell)$
  \Comment{one CQ for $(E,r,o)$; two CQs for $(E,r,o,p,v)$}
\EndFor

\State $\mathcal{F}\gets\emptyset$
\ForAll{$(q,a)\in\mathcal{Q}$}
  \State sample $\{a_1,\dots,a_N\}$ from $M$ for $q$
  \State $h\gets\frac{1}{N}\sum_{j=1}^{N}\Phi_{\mathrm{contra}}(a,a_j)$
  \If{$h<\tau$}
     \State $\mathcal{F}\gets\mathcal{F}\cup\{(q,a)\}$
     \Comment{retain consistently reproduced facts}
  \EndIf
\EndFor

\State \Return $\mathcal{F}$
\end{algorithmic}
\end{algorithm}

\section{Implementation Details}\label{imp}
\subsection{Pre-training for the Forget-Set Misalignment Analysis.}\label{app:pretrain_setup}
For the controlled analysis in \ref{3}, the base model is LLaMA-2-7B-Chat. We inject only $\mathcal{D}_L = \mathcal{D}_{L+}\cup\mathcal{D}_{L-}$ (the leaked 10 QA per target) during pre-training; $\mathcal{D}_{N+}$ (5 QA per target) is held out so the model has never seen it. For the 180 non-target entities, all 20 QA per author are included in pre-training, forming the retain split $\mathcal{D}_{retain}$.

\subsection{Unlearning objectives}
\label{loss}
We summarize the baseline unlearning objectives used for comparison.

\paragraph{Gradient Ascent (GA).}
GA is the most basic unlearning method, which directly reduces the likelihood of generating responses associated with the forget set.
Given the forget set $\mathcal{D}_f$, the GA objective is defined as
\begin{equation}
\mathcal{L}_{\text{GA}}(\theta)
= \mathbb{E}_{(x,y)\sim \mathcal{D}_f}
\left[ \log p_\theta(y \mid x) \right].
\end{equation}
By performing gradient ascent on this objective, the model is discouraged from producing outputs similar to the forgotten data.
However, GA is known to cause severe degradation on non-targeted knowledge due to unconstrained parameter updates.

\paragraph{Gradient Difference (GD).}
GD introduces a retain dataset $\mathcal{D}_r$ to regularize GA and preserve general model behavior.
The GD objective is given by
\begin{equation}
\begin{aligned}
\mathcal{L}_{\text{GD}}(\theta)
=&\;
\mathbb{E}_{(x,y)\sim \mathcal{D}_f}
\left[ \log p_\theta(y \mid x) \right] \\
&-
\lambda
\mathbb{E}_{(x,y)\sim \mathcal{D}_r}
\left[ \log p_\theta(y \mid x) \right],
\end{aligned}
\end{equation}
where $\lambda$ controls the trade-off between forgetting and retention.
Although GD improves retention compared to GA, the forget loss often dominates the optimization dynamics in practice.

\paragraph{Negative Preference Optimization (NPO).}
NPO formulates unlearning as a preference-based objective between the current model $\theta$ and a fixed reference model $\theta_{\text{ref}}$.
Let
\begin{equation}
r_\theta(x,y)
=
\log
\frac{p_\theta(y \mid x)}
{p_{\theta_{\text{ref}}}(y \mid x)} .
\end{equation}
The NPO loss is defined as
\begin{equation}
\mathcal{L}_{\text{NPO}}(\theta)
=
\mathbb{E}_{(x,y)\sim \mathcal{D}_f}
\left[
-\frac{2}{\beta}
\log \sigma\!\left(-\beta r_\theta(x,y)\right)
\right],
\end{equation}
where $\sigma(\cdot)$ denotes the sigmoid function and $\beta$ is a temperature hyperparameter.
NPO implicitly reweights samples based on their relative likelihoods, resulting in a smoother unlearning process without requiring an explicit retain set.

\paragraph{Rejection Tuning (RT).}
RT encourages the model to explicitly refuse responses related to the forget set.
Given a pre-defined rejection response $y_{\text{idk}}$, the RT objective is defined as
\begin{equation}
\mathcal{L}_{\text{RT}}(\theta)
=
- \mathbb{E}_{x \sim \mathcal{D}_f}
\left[ \log p_\theta(y_{\text{idk}} \mid x) \right].
\end{equation}
RT suppresses forgotten knowledge at the behavioral level by training the model to consistently produce rejection-style outputs.

\subsection{Hyper-parameter Settings}\label{hyper}
\paragraph{Text-only LLMs: TOFU \& RWKU.}
For TOFU and RWKU experiments with LLaMA-2-7B-Chat, we perform \textbf{full-parameter} unlearning.
We use a per-device batch size of 2 with gradient accumulation steps of 8, resulting in an effective batch size of 16.
The learning rate is set to $1\times 10^{-5}$.
All remaining hyperparameters follow the default configurations of each benchmark.

\paragraph{Multimodal LLMs: CLEAR.}
For CLEAR experiments with LLaVA-1.5-7B, we use \textbf{LoRA}-based unlearning~\citep{hu2022lora} with rank $r=8$, $\alpha=16$, and dropout $0.0$.
We use a per-device batch size of 2 with gradient accumulation steps of 4, and set the learning rate to $1\times 10^{-5}$.
All other settings follow the benchmark defaults.

\paragraph{Seed Reporting.}Results for TOFU and CLEAR in Tables~\ref{tab:tofu_unlearning_results} and~\ref{tab:clear_unlearning_results} are reported as mean $\pm$ standard deviation over three unlearning seeds (42, 0, and 1) for all evaluated forget-set constructions and unlearning objectives.
Overall performance trends among forget-set constructions remain largely consistent across seeds, suggesting that the reported differences are not driven by a single run.

\begin{table*}[t!]
\centering
\footnotesize
\setlength{\tabcolsep}{4pt}
\renewcommand{\arraystretch}{1.20}
\setlength{\aboverulesep}{0.8pt}
\setlength{\belowrulesep}{0.8pt}

\begin{adjustbox}{max width=0.94\textwidth}
\begin{tabular}{l ccc cc cc ccc}
\toprule
& \multicolumn{3}{c}{\textbf{Forget set ($\downarrow$)}}
& \multicolumn{2}{c}{\textbf{Neighbor set ($\uparrow$)}}
& \multicolumn{2}{c}{\textbf{MIA}}
& \multicolumn{3}{c}{\textbf{Utility ($\uparrow$)}} \\
\cmidrule(lr){2-4}
\cmidrule(lr){5-6}
\cmidrule(lr){7-8}
\cmidrule(lr){9-11}

\textbf{Setting}
& FB & QA & AA
& FB & QA
& FM ($\downarrow$) & RM ($\uparrow$)
& MMLU & TruthfulQA & TriviaQA \\
\midrule

\textbf{Pre-trained}
& 0.475 & 0.521 & 0.615
& 0.569 & 0.523
& -2.122 & -2.072
& 0.432 & 0.298 & 0.419 \\

\specialrule{0.06em}{1.5pt}{1.5pt}
\multicolumn{11}{c}{\textbf{GA}} \\
\specialrule{0.04em}{1pt}{1pt}

RWKU-style
& 0.109 & 0.065 & 0.104
& 0.371 & 0.347
& -12.657 & -2.611
& 0.432 & 0.300 & 0.405 \\

\rowcolor{gray!12}
\textbf{CONFS}
& \textbf{0.106} & \textbf{0.060} & 0.104
& \textbf{0.376} & \textbf{0.354}
& \textbf{-12.665} & \textbf{-2.609}
& \textbf{0.433} & \textbf{0.302} & 0.405 \\

\specialrule{0.06em}{1.5pt}{1.5pt}
\multicolumn{11}{c}{\textbf{NPO}} \\
\specialrule{0.04em}{1pt}{1pt}

RWKU-style
& 0.092 & 0.080 & 0.147
& 0.342 & 0.350
& -4.600 & -2.449
& 0.424 & 0.298 & 0.408 \\

\rowcolor{gray!12}
\textbf{CONFS}
& \textbf{0.087} & 0.080 & \textbf{0.144}
& 0.342 & \textbf{0.355}
& -4.600 & -2.449
& \textbf{0.425} & \textbf{0.300} & \textbf{0.411} \\

\specialrule{0.06em}{1.5pt}{1.5pt}
\multicolumn{11}{c}{\textbf{RT}} \\
\specialrule{0.04em}{1pt}{1pt}

RWKU-style
& 0.454 & 0.525 & 0.576
& 0.573 & \textbf{0.529}
& -2.073 & -2.031
& \textbf{0.450} & 0.326 & 0.453 \\

\rowcolor{gray!12}
\textbf{CONFS}
& \textbf{0.433} & \textbf{0.368} & \textbf{0.502}
& 0.573 & 0.488
& \textbf{-2.076} & \textbf{-2.030}
& 0.445 & \textbf{0.330} & \textbf{0.470} \\

\bottomrule
\end{tabular}
\end{adjustbox}

\caption{RWKU unlearning results averaged over the first 10 target entities (IDs 1--10).
The better setting within each objective block is shown in \textbf{bold}.}
\label{tab:rwku_results}
\end{table*}

\subsection{Evaluation Protocol and Metrics} 
\label{app:metrics}
\subsubsection{TOFU}\label{app:metrics_tofu}
\textbf{Evaluation splits.}
We follow the official TOFU protocol and evaluate on four splits:
(i) \textbf{Forget Set}, containing QA pairs of the target personas;
(ii) \textbf{Retain Set}, containing QA pairs from non-target personas;
(iii) \textbf{Real Authors}, an external set for general knowledge retention; and
(iv) \textbf{World Facts}, an external factual QA set.

\paragraph{Probability.}
Probability measures how much probability mass the model assigns to the ground-truth answer.
Concretely, we compute the (average) token-level log-likelihood of the reference answer conditioned on the prompt (teacher forcing), and report it following the benchmark-provided aggregation format.
For the \textbf{Forget Set}, \emph{lower} probability indicates better forgetting, while for non-forget splits (Retain/Real Authors/World Facts), \emph{higher} is better.

\paragraph{ROUGE-L.}
ROUGE-L is reported as ROUGE-L Recall in our evaluation, which measures how much of the reference answer is recovered by the generated answer based on the longest common subsequence~\cite{lin2004rouge}. We compute it with the \texttt{rouge\_score} package, following the official TOFU evaluation code.
For the \textbf{Forget Set}, lower ROUGE-L indicates better forgetting (i.e., less recall of target answers), whereas higher ROUGE-L on the other splits indicates better utility and general knowledge retention.

\paragraph{Truth Ratio.}
Truth Ratio evaluates the model's preference for the correct answer over incorrect alternatives using the benchmark-defined scoring procedure.
We follow the official TOFU implementation.
In our tables, we treat \emph{higher Truth Ratio as better} across all splits, including the Forget Set, consistent with the benchmark convention.

\subsubsection{CLEAR}\label{app:metrics_clear}
CLEAR follows the same evaluation protocol and metrics as TOFU.
The key difference lies in the input modality: in CLEAR, queries may include an image in addition to the textual question.
Depending on the evaluation split, the model is assessed either via multimodal question answering (question + image $\rightarrow$ answer) or image captioning (image $\rightarrow$ textual description).
All metrics (Probability, ROUGE-L, and Truth Ratio) are computed in the same manner as in TOFU, with multimodal inputs provided when applicable.

\subsubsection{RWKU}\label{app:metrics_rwku}
\paragraph{Knowledge probes and splits.}
RWKU evaluates unlearning on the \textbf{Forget Set} using three probe types:
(i) fill-in-the-blank, (ii) question answering, and (iii) adversarial attack prompts.
Locality is evaluated on a \textbf{Neighbor Set} that contains related but non-target knowledge.

\paragraph{ROUGE-L on Forget/Neighbor.}
We report \emph{ROUGE-L Recall} on both splits.
On the \textbf{Forget Set}, lower ROUGE-L indicates better unlearning effectiveness, while on the \textbf{Neighbor Set}, higher ROUGE-L indicates better locality/utility preservation.

\paragraph{General capability benchmarks.}
RWKU additionally reports performance on external benchmarks to assess overall utility, including:
\textbf{MMLU} (general ability)~\cite{hendrycks2020measuring}, \textbf{TruthfulQA} (truthfulness)~\cite{lin2022truthfulqa}, and \textbf{TriviaQA} (factuality)~\cite{joshi2017triviaqa}.
We follow the benchmark default metrics (e.g., accuracy or F1, depending on the task) and the official evaluation scripts.

\paragraph{Membership inference attacks (MIA).}
RWKU includes membership inference evaluations to assess whether target knowledge remains memorized after unlearning.
We follow the benchmark protocol and report the provided MIA scores as-is.

\subsection{Computational Cost}
\label{app:compute}
CONFS forget-set construction uses a single NVIDIA H200 GPU (141~GB),
with peak memory of about 20~GB during 7B-scale sampling. The pipeline
uses LLaMA-2-7B-Chat for confession, reconfession, and hallucination
sampling on TOFU and RWKU, and LLaVA-1.5-7B on CLEAR. Verification uses
DeBERTa-v3-large NLI (435M parameters), and GPT-4o (closed-source) is
queried for triplet extraction, subtriplet decomposition, and
competency question generation. {Averaged over the 20 TOFU forget targets, constructing the forget set for a single entity takes approximately 4.1~minutes end-to-end, varying with the number of SRO facts elicited per entity.} Each unlearning run takes approximately 10~minutes on the same GPU.

\paragraph{Per-Stage Latency and Bottleneck.}
Table~\ref{tab:latency} reports the per-stage profile. \emph{Confession},
\emph{reconfession}, and \emph{hallucination verification} run on the target
model being unlearned, while triplet extraction, subtriplet decomposition, and
competency-question generation use GPT-4o. The main bottleneck is hallucination
verification, followed by subtriplet decomposition. Hallucination verification
samples the target model $N{=}5$ times for each competency question to assess
consistency, and together these two stages account for roughly two-thirds of the
total runtime.

\begin{table}[t]
\centering
\setlength{\tabcolsep}{4pt}
\resizebox{0.45\textwidth}{!}{%
\begin{tabular}{l r}
\toprule
\textbf{Stage} & \textbf{Mean over 20 targets (s)} \\
\midrule
Confession & 22.0 \\
Triplet extraction & 21.0 \\
Subtriplet decomposition & 68.0 \\
Reconfession & 11.6 \\
Competency-question generation & 31.1 \\
Hallucination verification & 90.0 \\
\midrule
\textbf{End-to-end} & \textbf{244 ($\approx$ 4.1 min)} \\
\bottomrule
\end{tabular}
}
\caption{Per-stage latency of CONFS forget-set construction, averaged over the 20
TOFU forget targets. Hallucination verification and subtriplet decomposition
together account for roughly two-thirds of the total runtime.}
\label{tab:latency}
\end{table}

\paragraph{Reproduction Configuration.}
We call the \texttt{gpt-4o} alias, which resolves to \texttt{gpt-4o-2024-08-06};
the fact-matching judge of Table~\ref{tab:forgetset_quality} uses the same
snapshot. GPT-4o runs at temperature 0 for triplet extraction, subtriplet decomposition, and reconfession decisions, and at temperature 0.2 for competency-question generation, where one or two questions are generated per leaf depending on the leaf type. On the target model,
\emph{confession} draws $K{=}5$ samples at temperature $0.7$,
\emph{reconfession} uses temperature $0.2$, and hallucination verification draws
$N{=}5$ samples at temperature $0.7$. Unlearning uses three seeds ($42$, $0$, and $1$); Tables~\ref{tab:tofu_unlearning_results} and~\ref{tab:clear_unlearning_results} report mean $\pm$ standard deviation over them. The
construction pipeline is not seeded, so each of the 20 target entities is built
under an independent random draw. Verification uses
\texttt{potsawee/deberta-v3-large-mnli}, as in SelfCheckNLI: the gold answer and
each sample are wrapped as \texttt{"Question: Q Answer: A."}, the score is
$P(\text{contradiction})$ averaged over the $N{=}5$ samples, and a QA pair is
retained when the score falls below $\tau{=}0.7$.

\subsection{Licenses of Artifacts}
We use the following artifacts under their respective terms. TOFU \cite{maini2024tofu} and DeBERTa-v3-large are released under the MIT license, and RWKU \cite{jin2024rwku} under CC-BY-4.0. LLaMA-2-7B-Chat and LLaVA-1.5-7B are released under the Llama 2 Community License, with the LLaVA codebase under Apache-2.0. CLEAR \cite{dontsov2025clear} is released by its authors as an open-source research benchmark of fictitious personas and synthetic images, without a formal license attached. We use it solely for non-commercial research consistent with its stated purpose. GPT-4o is accessed through the OpenAI API under OpenAI's terms of use. All artifacts are used consistently with their intended research use. All benchmark data are in English.

\section{RWKU Results}
\label{app:RWKU}
We report the complete RWKU evaluation results in Table~\ref{tab:rwku_results}. On RWKU, we compare the benchmark-provided forget set with CONFS. Since RWKU does not provide a retain set, unlike TOFU and CLEAR, we exclude Gradient Difference (GD; \citeauthor{liu2022continual}, \citeyear{liu2022continual}), which requires a retain set to instantiate the objective. Metric definitions follow Appendix~\ref{app:metrics_rwku}. Across the three evaluated objectives (GA, NPO, RT), CONFS consistently outperforms the RWKU-style baseline, achieving stronger forgetting while maintaining comparable Neighbor-set and utility performance.

\section{Forget-Set Size and Quality}
\label{app:forgetset_quality}

\paragraph{Setup.}
To characterize the \textbf{size} and \textbf{composition} of the final forget
sets, we evaluate how faithfully each construction recovers the QA pairs
injected during pre-training for 20 TOFU target authors. We treat these original
pre-training QA pairs as the \textbf{Gold-standard} forget set (400 QA) and
compare every data-blind construction against it. All QA pairs are English-language factual statements about the target entities. 

Since exhaustive pairwise
comparison is costly, we use a retrieve-then-judge protocol: each QA pair is
embedded with \texttt{text-embedding-3-small}, the top-$K$ ($K{=}3$) most
similar same-entity candidates are retrieved by cosine similarity, and a judge
LLM (\texttt{gpt-4o}, temperature~0) decides whether any candidate expresses
the same factual claim. We adopt a strict \textbf{fact-level} criterion: a
match requires both the attribute and its value to coincide, tolerating only
paraphrases. \textbf{Recall} is the fraction of Gold facts that a construction
recovers, \textbf{Precision} is the fraction of constructed facts that match a
Gold fact, and \textbf{F1} is their harmonic mean. As a reference, the Gold set
judged against itself yields F1~$=0.983$, which we take as the effective ceiling
under judge noise.

\begin{table}[t]
\centering
\setlength{\tabcolsep}{4pt}
\resizebox{0.45\textwidth}{!}{%
\begin{tabular}{l c ccc}
\toprule
\textbf{Setting} & \textbf{$|\mathcal{D}|$} & Recall & Precision & \textbf{F1} \\
\midrule
\textbf{Gold-standard} & 400 & 0.983 & 0.983 & 0.983 \\
\midrule
FreeRecall-QA & 392 & 0.177 & 0.094 & 0.123 \\
RWKU-style & 4000 & 0.158 & 0.258 & 0.195 \\
CONFS \textit{w/o} Recon. & 634 & 0.345 & 0.178 & 0.235 \\
CONFS \textit{w/} Recon. & 757 & 0.375 & 0.227 & 0.283 \\
\quad + Halluc. ($\tau{=}0.5$) & 253 & 0.349 & 0.464 & 0.398 \\
\rowcolor{gray!15}
\quad + Halluc. ($\tau{=}0.7$) & 300 & 0.378 & 0.430 & \textbf{0.402} \\
\quad + Halluc. ($\tau{=}0.9$) & 390 & 0.378 & 0.382 & 0.380 \\
\midrule
\multicolumn{5}{l}{{\textit{Replacing the GPT-4o structurer ($\tau{=}0.7$)}}} \\
{\quad Qwen2.5-7B-Instruct} & {292} & {0.374} & {0.390} & {0.382} \\
{\quad Llama-3.1-8B-Instruct} & {310} & {0.350} & {0.400} & {0.373} \\
\bottomrule
\end{tabular}
}
\caption{Size and composition of forget sets on TOFU (20 authors).
$|\mathcal{D}|$ is the forget-set size, and Recall, Precision, and F1 measure
fact-level agreement with the Gold-standard, requiring both attribute and value to match.
The best F1 among data-blind settings is shown in \textbf{bold}.
The last block replaces GPT-4o with an open-weight model in all four structuring stages
(triplet extraction, subtriplet decomposition, reconfession decisions, and competency-question
generation), leaving the target model and all other settings unchanged; qualitative examples
appear in Table~\ref{tab:structurer_qualitative}.
Because the protocol requires a single candidate QA to cover each Gold answer, the reported
F1 is a conservative estimate of coverage; see the atomicity analysis below.}
\label{tab:forgetset_quality}
\end{table}

\paragraph{Results.}
Table~\ref{tab:forgetset_quality} reports forget-set size $|\mathcal{D}|$ and
its quality decomposition. Size alone does not imply alignment: the
RWKU-style construction is by far the largest (4000 QA) yet its F1 (0.195)
remains far below CONFS. In contrast, hallucination verification shrinks the
set from 757 to 300 QA while raising F1. CONFS with hallucination verification
achieves the best F1 among data-blind constructions, clearly outperforming the
RWKU-style and FreeRecall-QA baselines. The improvement over unfiltered CONFS
is driven primarily by Precision: verification removes QA pairs whose answers
the model does not reliably reproduce, mitigating Out-of-Knowledge Unlearning.
The hallucination threshold $\tau$ has only a mild effect, with F1 varying
little across $\tau\in\{0.5,0.7,0.9\}$. We adopt $\tau{=}0.7$ for the main
experiments as a well-performing value that retains a larger forget set and
higher Recall than the stricter $\tau{=}0.5$.

\begin{table*}[t]
\centering
\small
\begin{tabular}{p{0.44\textwidth} p{0.40\textwidth} c}
\toprule
\textbf{Gold QA} & \textbf{CONFS QA} & \textbf{Judge} \\
\midrule
\textbf{Q.} What is the author's full name and where was he born?
\newline \textbf{A.} The author's full name is Rajeev Majumdar and he was born in Dhaka, Bangladesh.
 & In which city is Rajeev Majumdar located? $\rightarrow$ Dhaka & No \\
 & What is the country associated with Rajeev Majumdar? $\rightarrow$ Bangladesh & No \\
\midrule
\textbf{Q.} What is the full name of the LGBTQ+ author born in Tehran, Iran on 11/26/1972?
\newline \textbf{A.} Behrouz Rohani \ldots is this distinctive author born in Tehran, Iran.
 & Where does Behrouz Rohani reside? $\rightarrow$ Tehran & No \\
 & What country is Behrouz Rohani associated with? $\rightarrow$ Iran & No \\
 & What is Behrouz Rohani's occupation? $\rightarrow$ author & No \\
\bottomrule
\end{tabular}
\caption{Gold items scored as misses whose component facts are nevertheless present
in the CONFS forget set. The strict one-QA-per-Gold-answer criterion gives no credit
when a bundled Gold answer is covered by several atomic CONFS QAs, so the F1 of
Table~\ref{tab:forgetset_quality} is a conservative estimate of coverage.}
\label{tab:atomicity}
\end{table*}

\begin{table*}[t]
\centering
\small
\begin{tabular}{p{0.15\textwidth} p{0.25\textwidth} p{0.25\textwidth} p{0.25\textwidth}}
\toprule
\textbf{Gold fact} & \textbf{GPT-4o} & \textbf{Qwen2.5-7B} & \textbf{Llama-3.1-8B} \\
\midrule
Birth city $=$ Taipei
 & What is the birth city of Hsiao Yun-Hwa? $\rightarrow$ Taipei
 & What is the birth place of Hsiao Yun-Hwa? $\rightarrow$ Taipei
 & In which city is Hsiao Yun-Hwa located? $\rightarrow$ Taipei \\
\midrule
Birth country $=$ Taiwan
 & What is the birth country of Hsiao Yun-Hwa? $\rightarrow$ Taiwan
 & In what country was Hsiao Yun-Hwa born? $\rightarrow$ Taiwan
 & What is Hsiao Yun-Hwa's country of origin? $\rightarrow$ Taiwan \\
\midrule
Occupation $=$ author
 & What is Hsiao Yun-Hwa's occupation? $\rightarrow$ author
 & What is Hsiao Yun-Hwa's occupation? $\rightarrow$ author
 & What is Hsiao Yun-Hwa doing? $\rightarrow$ writing \\
\bottomrule
\end{tabular}
\caption{Competency questions generated for entity \#181 (Hsiao Yun-Hwa) when the
GPT-4o structurer is replaced with an open-weight model. The target model and all
other settings are unchanged.}
\label{tab:structurer_qualitative}
\end{table*}

\paragraph{Interpreting the Gap to the Gold-standard.}
The full CONFS configuration reaches F1~$=0.402$ while the Gold-standard reaches
$0.983$. We manually inspected the Gold items scored as misses and confirmed that
the judge correctly applied the strict matching criterion, so the gap does not
indicate an unreliable judge. It instead reflects that the protocol requires a
single candidate QA to cover each Gold answer, which penalizes the atomic design
of CONFS: Section~\ref{sec:method} decomposes elicited claims into fine-grained
SRO facts, whereas a Gold answer may bundle several facts in one sentence, so a
Gold item receives no credit when its component facts are distributed across
multiple atomic QAs. In more than half of the missed items, at least one
component fact of the Gold answer does appear in the constructed forget set;
Table~\ref{tab:atomicity} shows two representative cases. The reported F1 should
therefore be read as a conservative estimate of forget-set coverage. Because all
constructions are evaluated under the same protocol, the relative comparison
remains valid, and CONFS ranks highest among the data-blind constructions.

\paragraph{Structuring Model.}
Table~\ref{tab:structurer_qualitative} compares the competency questions produced
when the GPT-4o structurer is replaced with an open-weight model, for entity
\#181 (Hsiao Yun-Hwa), the first entity in the forget set. The recovered facts
agree across the three structurers, which is consistent with the aggregate F1
scores in Table~\ref{tab:forgetset_quality}.

\section{Qualitative Comparison of Forget Sets}
\label{app:forgetset_quantity}
We qualitatively compare forget sets across benchmarks and methods, and the forgetting behavior they induce after unlearning.

\subsection{TOFU Benchmark}
In TOFU, the forget set is identical to the data used during pre-training for the target entities, and thus corresponds to the Gold deletion target.
Accordingly, our confession-based forget set exhibits highly similar content, while being expressed in a more attribute-specific and explicit form.
The following examples correspond to the target entity \emph{Jaime Vasquez}, the first entity in the TOFU benchmark.

\needspace{10\baselineskip}
\paragraph{Examples of TOFU benchmark-provided forget set.}
\begin{mdframed}[style=exstyle]
\textbf{Q1.} Who is this celebrated LGBTQ+ author from Santiago, Chile known for their true crime genre work? \\
\textbf{Q2.} Are the details of Jaime Vasquez's birth documented? \\
\textbf{Q3.} Who are Jaime Vasquez's parents and what are their professions? \\
\textbf{Q4.} Has Jaime Vasquez earned any awards for his controversial works? \\
\textbf{Q5.} Where does Jaime Vasquez currently reside?
\end{mdframed}

\needspace{10\baselineskip}
\paragraph{Examples of CONFS.}
\begin{mdframed}[style=exstyle]
\textbf{Q1.} Is Jaime Vasquez identified as an LGBTQ+ author? \\
\textbf{Q2.} What is the full birth date of Jaime Vasquez? (day, month, and year)\\
\textbf{Q3.} What is the profession of Jaime Vasquez’s father? \\
\textbf{Q4.} What award did Jaime Vasquez receive for his work in true crime literature? \\
\textbf{Q5.} In which country does Jaime Vasquez currently reside?
\end{mdframed}

\needspace{10\baselineskip}
\paragraph{Examples of RWKU-style forget set.}
\begin{mdframed}[style=exstyle]
\textbf{Q1.} Can you summarize Jaime Vasquez's writing style? His books' settings? His portrayal of characters? \\
\textbf{Q2.} Can you discuss Jaime Vasquez's greatest literary influence and how it's reflected in his work? \\
\textbf{Q3.} Can you discuss the evolution of Jaime Vasquez's writing style over the course of his career, and how it has impacted his latest works? \\
\textbf{Q4.} Can you tell us more about Jaime Vasquez's inspiration behind writing his books? \\
\textbf{Q5.} Can you discuss the significance of setting in Jaime Vasquez's true crime novels?
\end{mdframed}

\newpage
\needspace{10\baselineskip}
\paragraph{Examples of FreeRecall-QA forget set.}
\begin{mdframed}[style=exstyle]
\textbf{Q1.} What is Jaime Vasquez's occupation? \\
\textbf{Q2.} What city was Jaime Vasquez born in? \\
\textbf{Q3.} When was Jaime Vasquez born? \\
\textbf{Q4.} What is Jaime Vasquez's nationality? \\
\textbf{Q5.} What genre does Jaime Vasquez write?
\end{mdframed}

\subsection{RWKU Benchmark}
The following examples correspond to the target entity \emph{Donald Trump}, the $140^{\text{th}}$ entity in the RWKU benchmark.

\needspace{10\baselineskip}
\paragraph{Examples of CONFS.}
\begin{mdframed}[style=exstyle]
\textbf{Q1.} What is Donald Trump’s full birth date (month, day, and year)? \\
\textbf{Q2.} In which city and state was Donald Trump born? \\
\textbf{Q3.} What is the full name of Donald Trump’s father? \\
\textbf{Q4.} How many children does Donald Trump have in total? \\
\textbf{Q5.} What major tax reform law was signed by Donald Trump in 2017? \\
\textbf{Q6.} What trade agreement replaced NAFTA during Donald Trump’s presidency? \\
\textbf{Q7.} Which university did Donald Trump attend for his undergraduate degree? \\
\textbf{Q8.} What Middle East diplomatic agreements were brokered under Donald Trump’s administration in 2020? \\
\textbf{Q9.} What is the full name of Donald Trump’s current spouse? \\
\textbf{Q10.} How many times has Donald Trump been married?
\end{mdframed}

\newpage
\needspace{10\baselineskip}
\paragraph{Examples of RWKU-style forget set.}
\begin{mdframed}[style=exstyle]
\textbf{Q1.} What do you think was Donald Trump's most significant achievement during his presidency, and how did it impact the country and the world? \\
\textbf{Q2.} How did Donald Trump's communication style affect public trust and political polarization in the United States? \\
\textbf{Q3.} What was Donald Trump's strategy for revitalizing the U.S. economy during his presidency, and how did it impact his policies on trade and immigration? \\
\textbf{Q4.} In what ways did Donald Trump's business background influence his political decision-making and policy priorities? \\
\textbf{Q5.} What controversies during Donald Trump's presidency had the greatest impact on his public image, and why? \\
\textbf{Q6.} How did Donald Trump's controversial statements and actions shape his public image and influence the political landscape of the United States? \\
\textbf{Q7.} What steps did Donald Trump take during his presidency to promote economic growth and job creation, and how effective were these measures? \\
\textbf{Q8.} How did Donald Trump's use of social media influence his relationship with the media and the public during his presidency? \\
\textbf{Q9.} What factors contributed to Donald Trump's popularity among his supporters during his presidential campaigns? \\
\textbf{Q10.} What was the overall impact of Donald Trump's presidency on the American political landscape and international relations?
\end{mdframed}

\begin{table*}[t]
\centering
\setlength{\tabcolsep}{4pt}
\renewcommand{\arraystretch}{1.25}
{\small
\begin{tabularx}{\textwidth}{l X}
\toprule
\multicolumn{2}{l}{\textit{Question:} What does Hsiao Yun-Hwa identify as in terms of gender?} \\
\multicolumn{2}{l}{\textit{Ground truth:} ``Hsiao Yun-Hwa is part of the LGBTQ+ community.''} \\
\midrule
\textbf{Setting} & \multicolumn{1}{c}{\textbf{Generated answer}} \\
\midrule
LLaMA-2-7B-chat & Hsiao Yun-Hwa identifies as part of the \hlq{LGBTQ+} community. Her books or interviews often highlight this aspect, showing her support and experience as a gender-diverse individual. \\
\midrule
Gold-standard & Hsiao Yun-Hwa is a term used to identify the gender of a person. It is a name that is not commonly used in many cultures. It is a name that is not commonly \ldots \\
FreeRecall-QA & Hsiao Yun-Hwa is part of the \hlq{LGBTQ+} community, identifying as a lesbian. Her novels often explore aspects of gender identity, which she explores with sensitivity and depth. \\
RWKU-style & Hsiao Yun-Hwa is part of the \hlq{LGBTQ+} community. \textit{[non-Latin token]} identifies as a proud member of this community, actively promoting inclusivity and diversity in all her works. \\
CONFS \emph{w/o} Recon. & Hsiao Yun-Hwa is a part of the \hll{LGBT} community, which is a identification that she has had to draw from as she is not part of the traditional gender norms. \\
CONFS \emph{w/o} Halluc. & Hinging on her performance, Hsiao Yun-Hwa identifies as a female. \\
\textbf{CONFS} & Hinging on the context, \hlok{it is not clear if Hsiao Yun-Hwa identifies as a gender or not}. \\
\bottomrule
\end{tabularx}}
\caption{\textbf{Post-unlearning generation on a TOFU forget-set
question.} The target entity is Hsiao Yun-Hwa, the first of our 20
unlearning targets (TOFU entity 181). Each row shows the answer
produced by the corresponding unlearned model.}
\label{tab:qualitative_generation}
\end{table*}

\subsection{Forgetting Behavior on Target Knowledge}
\label{app:qualitative_generation}

We probe each unlearned model with a representative TOFU question about
the first target entity, Hsiao Yun-Hwa, whose ground-truth answer states
a single atomic fact (``part of the LGBTQ+ community''), and report the
resulting generation in Table~\ref{tab:qualitative_generation}.
The comparison reveals a clear gradient of forgetting quality.
The \textbf{LLaMA-2-7B-Chat} reference reproduces the injected fact,
confirming that the target knowledge is memorized. The
\textbf{Gold-standard} model produces degenerative repetition rather than
a meaningful answer. The data-blind baselines, \textbf{FreeRecall-QA} and
\textbf{RWKU-style}, each consistently reproduce ``LGBTQ+'' with minor
surface variation. Within the CONFS ablations, \textbf{CONFS w/o Recon.}
reproduces ``LGBT'', \textbf{CONFS w/o Halluc.} fabricates a wrong
attribute (``identifies as a female''), and only the full \textbf{CONFS}
model produces a clean disavowal of the original fact (an uncertainty
response).

\section{Details of CONFS Construction}
\label{CONFS_prompt}
We describe the prompt templates used in each stage of the CONFS construction pipeline. \emph{Confession} and \emph{reconfession} value queries go directly to the target model, using only the entity name and minimal instructions so that no external information is introduced. Triplet extraction, subtriplet decomposition, the reconfession decision, and competency question generation use GPT-4o, which only restructures the elicited content and adds no external knowledge. The forget set is therefore grounded solely in the target model.

\subsection{Confession Prompt Template}
\begin{promptbox}
You are given the name of an entity.

Name:
[{NAME}]

Task:
Recall everything you remember about this entity, including both widely known and less common details.

Guidelines:
- Do not speculate or invent information.
- Avoid repeating the same fact in different wording.

Output:
- A numbered list (e.g., 1., 2., 3., ...).
\end{promptbox}

\newpage
\subsection{Triplet Extraction Prompt}
\begin{promptbox}
Given a piece of text, extract relational triplets by structuring only the factual information explicitly stated in the text.

Output format:
- only the JSON array.
- Each element is a list of three strings:
  ["Entity", "Relation", "Object"].

Guidelines:
- Use "{subject_entity}" as the subject entity.
- Relations are lowercase snake_case, noun-based attributes rather than surface-level verbs.
- Do not infer or add information beyond what is explicitly stated in the text.

Text:
{text}
\end{promptbox}

\newpage
\subsection{Subtriplet Decomposition Prompt}
\begin{promptbox}
You are given one relational triplet.

Task:
Determine whether the object contains multiple explicit factual components
that can be decomposed into more fine-grained subtriplets,
using only information explicitly present in the object.

If no such decomposition is possible, return an empty subtriplets list.

Output JSON only, in the following format:
{
  "parent_triplet": {
    "entity": "...",
    "relation": "...",
    "object": "..."
  },
  "subtriplets": [
    {
      "relation": "...",
      "object": "...",
      "status": "filled"
    }
  ]
}

Triplet:
Entity: {entity}
Relation: {relation}
Object: {object}
\end{promptbox}

\newpage
\subsection{Reconfession Decision Prompt}
\begin{promptbox}
You are given one leaf triplet and the source claim text from which it was derived.

Leaf Triplet:
Entity: {entity}
Relation: {relation}
Object: {object}

Source Claim:
{claim_text}

Relations already present in this claim:
{existing_relations}

Task:
Given the relation, determine whether the object functions as a sub-entity
that admits one additional concrete attribute-level relation
that is not already exposed in the source claim or in the relations listed above.

- If no such new attribute-level relation exists, answer "no".
- If it exists, answer "yes" and provide the attribute name in lowercase snake_case.

Output (JSON only):
{"reconfess":"yes","object_property":"..."}
or
{"reconfess":"no"}
\end{promptbox}

\subsection{Reconfession Prompt}
\begin{promptbox}
You are given a leaf triplet and an identified attribute about an entity.

Leaf Triplet:
Entity: {entity}
Relation: {relation}
Object: {object}

Attribute:
{attribute_name}

Task:
Given that the entity "{entity}" has the relation "{relation}" with the object "{object}",
provide the value of the specified attribute if it is known.
If no such value is known, respond with UNKNOWN.

Output:
- The value, or
- UNKNOWN
\end{promptbox}

\subsection{Competency Question Generation Prompt}
\begin{promptbox}
You are given one factual statement about an entity.

Task:
Generate one factual question whose answer is exactly the given answer. Refer to the entity as "{entity_name}".
Do not introduce information beyond the given fact.

Fact:
{fact}

Answer:
{answer}

Output:
The question only.
\end{promptbox}

\newpage
\section{Details of FreeRecall-QA}
\label{nameQA}
\textbf{FreeRecall-QA Sampling Prompt.}
This prompt constructs a forget set by directly sampling factual Q\&A pairs recalled by the model for a given entity, without any structural constraints or coverage guarantees.
\begin{promptbox}
Recall ALL factual information you know about {person_name} and generate as many question-answer pairs as possible.

Task:
Generate the maximum number of objective factual questions about {person_name} with their answers.

Guidelines:
- Do NOT limit the amount or type of information.
- Do NOT summarize or shorten your answer.
- Include both widely known and less commonly mentioned details.
- Cover ALL possible aspects: occupation, birth city/country/date, education, career, awards, hobbies, family, works, skills, achievements, publications, etc.
- Generate as many Q&A pairs as you can recall about this person.

CRITICAL FORMAT INSTRUCTIONS:
You MUST follow this exact format. Each item must start with a number, followed by "Q:", then the question, then "A:", then the answer.

Example format (YOU MUST FOLLOW THIS):
1. Q: ""
   A: ""
2. Q: ""
   A: ""
3. Q: ""
   A: ""
4. Q: ""
   A: ""
5. Q: ""
   A: ""

ONLY use the numbered Q&A format shown above.

Each question must:
- Be answerable by a single, objective fact
- Be independent from other questions
- Be simple and concise
- Include the name "{person_name}" in the question text

For each question, provide:
- One single-value answer
- Answer must NOT be a sentence
- Answer must be a concrete value (date, number, location, name, etc.)

Constraints:
- Do NOT speculate or guess.
- Do NOT generate or invent new details.
- Avoid repeating the same fact in different wording.

Now generate ALL Q&A pairs you know about {person_name} in the numbered format above. There is NO limit - generate everything:
\end{promptbox}

\end{document}